  \documentclass[final,5p,times,twocolumn]{elsarticle}

\usepackage{amssymb}
\usepackage{url}
\usepackage{hyperref}
\usepackage{placeins}
\usepackage{graphicx}
\usepackage{tcolorbox}
\usepackage{subcaption}
\usepackage{amsmath}
\usepackage{algorithm}
\usepackage{tablefootnote}
\usepackage{multicol}
\usepackage{caption}
\usepackage{algorithmic}
\usepackage{colortbl}
\usepackage{xcolor}
\usepackage{longtable}
\usepackage{booktabs}

\usepackage{booktabs}
\usepackage{array}
\usepackage{url}
\usepackage{booktabs}
\usepackage{multirow}
\UseRawInputEncoding

\FloatBarrier
\journal{.}

\begin{document}

\begin{frontmatter}



\title{Global Ease of Living Index: A Machine Learning Framework  for Longitudinal Analysis of Major Economies}


 \author[inst1,inst2]{Arun Kumar Selvaraj }
\author[inst1,inst2]{Tanay Panat}
\author[inst1]{Rohitash Chandra}

\affiliation[inst1]{organization={Transitional Artificial Intelligence Research Group, School of Mathematics and Statistics},
            addressline={University of New South Wales}, 
            city={Sydney},
            postcode={2052}, 
            state={NSW},
            country={Australia}}

\affiliation[inst2]{organization={Centre for Artificial Intelligence and Innovation}, 
            addressline={Pingla Institute}, 
            city={Sydney}, 
            country={Australia}}

\begin{abstract} 

The drastic changes in the global economy, geopolitical conditions, and disruptions such as the COVID-19 pandemic have impacted the cost of living and quality of life. It is essential to comprehend the long-term implications of the cost of living and quality of life in major economies. A transparent and comprehensive living index must include multiple dimensions of living conditions.
In this study, we present an approach to quantifying the quality of life through the  Global Ease of Living Index that combines various socio-economic and infrastructural factors into a single composite score. Our index utilises economic indicators that define living standards, which could help in targeted interventions to improve specific areas. We present a machine learning framework to address missing data for certain economic indicators in specific countries. We then curate and update the data and use a dimensionality reduction approach (Principal Component Analysis and Factor Analysis) to create the Ease of Living Index for major economies since 1970. Our work significantly adds to the literature by offering a practical tool for policymakers to identify areas needing improvement, such as healthcare systems, employment opportunities, and public safety. Our approach with open data and code can be easily reproduced and applied to various contexts, providing transparency and accessibility for ongoing research and policy development in quality-of-life assessment.

\end{abstract}

\begin{keyword}
 Quality of Life\sep Socio-Economic Factors\sep Composite Score\sep Living Conditions\sep Data transparency\sep Reproducibility 
\end{keyword}

\end{frontmatter}


\section{Introduction}
\label{sec:sample1}
The assessment of the quality of life has evolved significantly over the years, incorporating a wide range of socioeconomic and infrastructural factors \cite{multifac_index}. Quantitative indexes, such as the World Happiness Index, rank countries based on various quality of life factors\cite{helliwell2012world}. Measuring happiness and quality of life is complex; current methods for assessing happiness might not be directly comparable, thus questioning the validity of averaging happiness and quality of life \cite{Ng}. 

There has been a lot of criticism towards existing indexes for their biases towards subjective measures. Research has shown that the  GDP per capita can significantly and negatively impact the happiness index \cite{Impact_macro_factors}, and quality of life index \cite{Quality_dependence}, suggesting that economic metrics heavily influence reported happiness levels \cite{supa}. Factors such as leisure, inequality, mortality, morbidity, crime, and the natural environment play significant roles in determining living standards but are not adequately reflected in GDP measurements \cite{Jones}. In addition to these challenges, the multidimensional nature of well-being \cite{well_being} necessitates a more comprehensive approach. Previous studies have often isolated economic factors, neglecting the interplay between socio-economic conditions and infrastructural elements. This oversight limits the understanding of how diverse factors collectively influence living standards \cite{Zanon}.

Several major indices for ranking countries in the domain of quality of life and ease of living exist, but they have major limitations. The Global Livability Index \cite{Livebility_Index} primarily focuses on stability, healthcare, culture, and environment, but it frequently lacks detailed data on social cohesion and community engagement, which are vital for understanding the quality of life in urban areas. Moreover, the Quality of Life Index \cite{Quality_life_index_gaps}, while encompassing various dimensions of urban living, often relies on aggregated data that may not reflect local realities. This is particularly evident in regions where data collection mechanisms are underdeveloped, leading to significant gaps in information on critical aspects such as public safety \cite{yu2014assessing}, environmental sustainability \cite{environmental_sustainability_gaps},  and economic inclusiveness. 

The World Happiness Report \cite{World_Happiness_Report} utilises data from surveys that ask people to respond to evaluate their current lives on a scale, often referred to as the "happiness ladder". A notable flaw in this index is its reliance on self-reported data, which can be influenced by cultural biases and social desirability, leading to discrepancies between reported happiness and actual well-being \cite{Flaw_HI}. The Happiness Index developed by the Happiness Alliance employs a comprehensive survey instrument that assesses multiple dimensions of well-being, yet it does not explicitly address how each dimension is weighted in the overall index calculation\cite{HI_Issue}. This lack of clarity can lead to criticisms regarding the transparency and validity of the index. Another critical issue is the potential for bias in the data collection process. The happiness index is calculated using a combination of health, economic, and social indicators. It often includes metrics such as life expectancy, education level, and income. Quality of Life \cite{QOL_Issues} assessments rely on self-reported data, which can introduce self-selection bias and affect the generalisability of the findings. There is also a predominant issue in the selection of dimensions and indicators that would truly represent a dynamic Quality of Life Index \cite{QOLI_EU_Issue}. The European Quality of Life Index, for instance, attempts to address this complexity by incorporating a wide range of indicators, but this can also lead to difficulties in ensuring that all relevant aspects of quality of life are adequately represented\cite{EU_QOLI}.

The Indian Ministry of Housing and Urban Affairs (MoHUA) launched the Ease of Living Index (EoLI) \cite{mohua2018ease} in 2018, which serves as a comprehensive tool to measure the quality of life in Indian cities. The EoLI is designed to assess the performance of 111 cities across institutional, social, economic, and physical pillars. These pillars are further divided into 15 categories, incorporating 78 indicators, of which 56 are core indicators and 22 are supporting. The institutional pillar evaluates governance, public services, and participatory governance. The social pillar focuses on key societal aspects such as education, health, identity, culture, and safety. The economic pillar assesses growth potential, employment opportunities, and inclusiveness of economic development, while the physical pillar evaluates city infrastructure, including housing, transportation, waste management, and environmental sustainability \cite{mohua2018ease}. The EoLI employs a robust model that can be adapted to various global settings.
Additionally, the application of the EoLI framework globally could facilitate the identification of best practices and lessons learned from different urban contexts. For example, countries facing rapid urbanisation \cite{zhang2016trends}, such as those in Southeast Asia, could benefit from the EoLI's structured approach to assessing infrastructure needs and social services \cite{Global_implementation_ELOI}.

The Cost of Living Index \cite{abraham2003toward}, the Pollution Index \cite{Junger_Leon}, the Healthcare Index \cite{kabir2022non}, the Crime Index \cite{targonski2012ucr} and Carbon-dioxide Emission \cite{macknick2009energy} are examples of indices that suffer from incomplete datasets, particularly in developing nations where data collection mechanisms may be underdeveloped\cite{Missing_imp_index_values}. The absence of indicators related to social aspects, such as community engagement and cultural identity, further limits the effectiveness of these indices in capturing the full spectrum of urban living conditions \cite{Missing_imp_index_values}. Data imputation is a crucial technique for addressing the challenges posed by missing data \cite{chen2024deep}. Machine learning models such as random forests and neural networks have been prominent for data imputation \cite{RFR_2}, including time series and tabular datasets.

Dimensional reduction methods such as Principal Component Analysis (PCA) and Factor Analysis can be utilised for index creation\cite{PCA_Index}.  Muratov \cite{Index_FA_Agri} employed factor analysis to evaluate the impact of innovations on farm productivity and construct an index reflecting the use of innovative practices in agricultural activities. Factor Analysis has also been deployed to create ecological factors such as the sustainable development index for the river basin in Ronda, Brazil \cite{FA_Ecological_Index} and integrate environmental, economic, social, and institutional performance indicators\cite{FA_Ecological_Index}. It has also been applied to develop a comprehensive evaluation index that captures various dimensions of educational performance \cite{Edu_FA}. This helps illustrate how factor analysis can help narrow the gap in educational attainment across different regions by providing a structured framework for evaluating and comparing educational institutions \cite{Edu_FA}. Similarly, PCA  has been used to create a Multidimensional Poverty Index (MPI) \cite{Poverty_Index_PCA} and analyse various poverty-related indicators, allowing researchers to derive weights that reflect the relative importance of each indicator in contributing to the overall poverty measurement. PCA has also been used to construct economic indices,  such as the cryptocurrency index \cite{shah2021principal}, where PCA normalised the data to construct the cryptocurrency index by aggregating the selected principal components based on their explained variance, ensuring that more significant components have a larger influence on the final index. Wang et al. \cite{wang2022sgl}  presented a  modified PCA for healthcare index creation from high-dimensional sensor data that incorporates sparsity and group lasso techniques to enhance the interpretability of the principal components.

In this study, we present a framework for constructing a global  Ease of Living Index (Global-EoLI) utilising machine learning for data imputation and index weightage computation. We evaluate the various factors, such as health and economic indicators, and weight assignments, to compute the final Global-EoLI from its underlying sub-indices. Using factor analysis to identify the most significant components contributing to each dimension, we compute the sub-indices such as economic, institutional, quality of life, and physical infrastructure.  We compare different methods for calculating and validating the index based on the sub-indices. The sub-indices are then assigned different weights to reflect their relative importance before aggregating them into the comprehensive Global-EoLI.
We utilise key machine learning approaches to effectively handle missing values based on the patterns in the observed data. We compare two methods based on their ability to maintain the variability and correlation structure of the dataset, ensuring robust imputation results.
The dataset used in this study spans the period from 1970 to 2022, covering a wide range of socioeconomic indicators for all available countries. However, our primary goal is the creation of the Global-EoLI for countries such as the United States, China, India, and Australia, while also examining global trends of G20 countries. We aim to deliver a reliable and meaningful index that captures the multidimensional aspects of living standards across countries, offering insights into temporal and regional patterns.

The rest of the paper is presented as follows. Section 2 reviews related indices, highlighting their methodologies and limitations. Section 3 presents the methodology covering data collection and imputation methods. Section 4 presents the results and analysis, comparing the index across different regions. Section 5 provides a discussion, and Section 6 provides the conclusions.

\section{Related Work}

\subsection{Machine Learning for Data Imputation }

Machine learning has emerged as a powerful tool for addressing the challenge of missing data imputation, particularly in time series datasets. Traditional imputation methods often struggle with the complexities and dependencies inherent in such data \cite{chen2024deep}.  In recent years, Generative Adversarial Networks (GANs) \cite{GAN} have gained prominence due to their ability to model complex data distributions. An implementation of GAN, the Generative Adversarial Imputation Network (GAIN) has been shown to outperform conventional methods by effectively capturing the underlying data structure and generating plausible imputations for missing values in various datasets, including clinical and environmental data\cite{Dong_GAIN,Yang_GAIN}.  In addition to GANs, other machine learning models have also been explored for imputation tasks. For example, Random Forests have been combined with generative models \cite{GAN_RF} to enhance the robustness of imputations, particularly in scenarios with large gaps in data. Recent developments in deep learning have introduced models such as Bidirectional Recurrent Imputation for Time Series (BRITS) \cite{BRITS}, which specifically address the temporal dependencies in time series data. BRITS employ a recurrent neural network (RNNs) \cite{yu2019review} architecture to handle multiple correlated missing values and adaptively update imputed values.
 
The data can be multifaceted, representing complex interactions between economic, institutional, social, and environmental dimensions. Due to the nature of our data, we require more sophisticated approaches to accurately capture and model these complexities, ensuring that the evolving trends in living standards are well-represented and analysed. Multiple Imputation by Chained Equations (MICE) \cite{MICE_imputation} is a widely adopted method for handling missing data in statistical analyses. This method allows us to account for the uncertainty associated with missing data by generating several plausible imputed datasets, which can then be analysed separately and combined to produce estimates that reflect this uncertainty.  MICE works under the assumption that data is missing completely at random \cite{little1988test}, this assumption may not be entirely appropriate as data can be missing due to underdeveloped data collection mechanisms in developing countries, due to variables related to governance such as crime or public safety may be systematically under-reported in regions where governments are less transparent and due to inconsistencies in data reporting due to several factors such as pandemics and wars.    


A recent study proposed a variational Bayesian deep learning framework for medical data imputation \cite{kulkarniChandra2024bayes}, allowing for a more nuanced understanding of the data's structure and dependencies using CATSI. This approach not only improves the accuracy of the imputations but also enhances the overall predictive performance of the models used for forecasting groundwater levels. Furthermore, the use of Bayesian inference, such as MCMC (Markov Chain Monte Carlo) sampling, has also been prominent in data imputation \cite{takahashi2017statistical}. Richardson et al. \cite{richardson2020mcflow}  presented a novel framework designed for data imputation that leverages the capabilities of Monte Carlo sampling and normalising flow generative models. They proposed an innovative iterative learning scheme that alternates between updating the density estimates of the data and imputing the missing values, thereby enhancing the robustness of the imputation process. Chen et al.  \cite{chen2024deep}  utilised MCMC sampling via a linear model for imputing data for groundwater modelling via deep learning models. 

Furthermore,  Denoising Autoencoder models (DAEs ) \cite{Chen_Shi} have shown promise in learning latent representations of data, which has been leveraged to impute missing values effectively \cite{Lall_Robin}. Gondara and Wang \cite{Gondara_Wang}  leveraged DAEs  to perform multiple imputations of missing data by reconstructing clean outputs from noisy inputs, which is particularly relevant for missing data scenarios where the absence of values can be viewed as a form of noise. Similarly, Pereira et al. \cite{Pereira_Santos} explored the potential of DAEs for missing data imputation, emphasising their capacity to learn from corrupted data. 
The flexibility of deep learning models allows them to adapt to different types of data and missing patterns, providing a significant advantage over traditional statistical methods.

Kulkarni and Chandra introduced Bayes-CATSI \cite{kulkarniChandra2024bayes}, a variational Bayesian approach tailored for medical time series data imputation that leveraged the probabilistic framework to project uncertainties in the data imputation. 
Data imputation via machine learning can be useful, particularly in index creation, where complete datasets are essential for accurate assessments. Another major challenge of creating an index is determining which factors are more important than others for determining ease of living and quality of life. After the data is acquired and imputed as needed, we need to decide based on the literature and expert knowledge on the weighting of the different sectors, such as economic, employment, crime and health indicators, to create an index. These processes are crucial in decision-making frameworks, especially in Multi-criteria Decision-Making (MCDM) \cite{omid}, where unequal weight assignment is often applied to reflect the varying importance of different criteria in real-world problems.  
\subsection{Machine learning for Index creation}

Machine learning has become an invaluable tool in developing and refining various indices that measure socio-economic factors \cite{keys2021machine}. Machine learning framework for  Doing Business Index \cite{DBI}  involves clustering countries and applying a multivariate Composite I-distance Indicator (CIDI) to derive data-driven weights for the index components. Furthermore, the application of machine learning techniques extends to other indices that assess economic performance and quality of life. For instance, a study demonstrated how machine learning can be utilised to predict stock market movements and stock price indices by employing trend deterministic data preparation methods\cite{Stock_Index}. This predictive capability is crucial for investors and policymakers alike, as it enables them to make informed decisions based on anticipated market trends. Machine learning techniques such as PCA have been employed to distil multiple indicators into a single composite score, allowing for a more nuanced understanding of different factors \cite{Reddy_reduction}.  PCA not only aids in reducing dimensionality but also enhances the interpretability of the data by highlighting the most significant variables \cite{Hiles_Dutcher}. The adaptability of PCA across various data types and its robustness against noise make it a preferred choice across fields, including social sciences\cite{Belay_Wondimu} and health studies\cite{Nowroozi_Roshani}.

 Tiwari et al.  \cite{Abtahi} presented the COVID-19 Vulnerability Index (C19VI), which integrated diverse data from public health databases, including demographic statistics and socioeconomic indicators. The study used machine learning models, specifically Random Forest and Support Vector Machines (SVM), for the identification of regions vulnerable to COVID-19. Furthermore,   Kheirati and Khan \cite{Khan_Ali} developed a Pavement Condition Index (PCI) using data from pavement condition surveys, incorporating visual inspections and surface distress measurements. The study applied machine learning models such as Decision Trees and Neural Networks, employing feature selection to enhance model accuracy and support effective maintenance planning. Keys et al. \cite{Isa} estimated the Human Footprint Index using ensemble learning techniques such as Gradient Boosting and Random Forests. The study integrated datasets on land use, population density, and ecological factors to capture the intricate relationships between human activities and environmental impacts, validated through spatial analysis techniques.

 \textcolor{black}{Our Global-EoLI employs variance-based weights derived from PCA and FA, which are the major indicators contributing more to the overall data variance and receive higher influence. This approach is data-driven, transparent and intended to capture dominant patterns in multidimensional living conditions.
In contrast, Bayesian methods estimate weights probabilistically using prior information and provide uncertainty quantification. Multidimensional Gini-based approaches assign weights based on inequality and dispersion across dimensions, focusing on distribution-sensitive aggregation rather than variance explained.
}

\section{Methodology}
\label{sec:sample2}

\subsection{Data}

In order to construct the Global-EoLI, we extracted data from multiple sources, including the International Monetary Fund (IMF), World Health Organisation (WHO),  and World Bank \cite{IMF2023,WHO2023,WorldBank2023}, and other sources as shown in the Appendix (Table \ref{tab:data_sources}). These data sources provided diverse socio-economic, healthcare, institutional, and quality-of-life indicators required to assess living standards across countries. 
 We reviewed existing indices, and the quality of life index can be seen as the closest to the Global-EoLI, with a major difference in ranking methodologies. The Global-EoLI employs data imputation for features with missing data and dimensionality reduction for the ranking process. 
We focus on data from 1970 to  2022  to develop the Global-EoLI and provide an assessment during this period.

\subsection{Data Imputation}

We encountered several challenges related to missing data in the data extraction process.  In the case of the cost of living index, the historical data was missing from 2012 to 1970, and the unemployment rate was missing from 1970 to 1990. The data for the gender development index was missing from 1970-1990, and we observed a pattern of \textit{missingness} for multiple indices as shown in Table \ref{tab:missingdata}. \textcolor{black}{We selected the features based on relevance, cross-country comparability, and availability across major economies.} Furthermore, imputing missing data using mean or mode values could misrepresent the underlying data patterns, and simply eliminating the rows with missing values would have resulted in a severe loss of information \cite{data_imputation_1}. 


Random Forests \cite{segal2004machine} employ an ensemble of decision trees to decide by voting or aggregation. The Random Forest Regressor (RFR) \cite{rf_3} is particularly effective in capturing complex relationships within the data, as it does not rely on the assumption of a linear relationship between variables.  Khan et al. \cite{RFR} reported that the RFR outperformed conventional data imputation methods, especially when dealing with complicated and high-dimensional data. García et al. \cite{RFR_2}  demonstrated that RFR improved imputation accuracy on multivariate time series datasets and efficiently filled in missing values by leveraging correlations between numerous variables. Further studies have shown that RFR outperforms traditional imputation methods \cite{RF_Imputer}, such as mean or median imputation, especially under conditions where data is missing completely at random (MCAR) or missing at random (MAR) \cite{RF_Imputer}. RFR implementation, in missForest \cite{missForest}, allows for the refinement of imputed values through repeated predictions until convergence is achieved \cite{Imputation_accuracy_RF}. This iterative method is better than single imputation techniques because it increases accuracy while lowering bias in the imputed values \cite{comparison_RF}.

MICE \cite{MICE_imputation} is a robust statistical technique that operates by creating multiple datasets with imputed values, allowing for errors in the imputation process to be included in later analyses. MICE is particularly advantageous for time series data, where missing values can disrupt the continuity and temporal relationships inherent in the data.  MICE  handles missing data by iteratively imputing each variable with missing values, conditioned on the other variables. MICE operates by updating these imputations over multiple cycles to ensure the imputed values reflect the underlying relationships in the data \textcolor{black}{as an alternative imputation strategy}.

\textcolor{black}{We selected Random Forest for imputation due to its ability to capture complex nonlinear relationships without requiring strong distributional assumptions, making it more robust and computationally efficient than   Bayesian methods that use MCMC sampling. MICE complement this by providing a statistically grounded approach that accounts for imputation uncertainty. We have not implemented methods such as  Generative Adversarial Networks (GAIN) \cite{yoon2018gain} and Bidirectional Recurrent Imputation (BRITS) \cite{cao2018brits} due to their computational cost and lower interpretability for tabular data.}

\textcolor{black}{In this study, the synthetic evaluation removed 40\% of observations from variables with relatively low missingness to assess imputation performance using RMSE and MAE. Although this approach allows comparison of imputation methods under controlled conditions, it does not fully replicate variables with extreme missingness (above 80 - 90\%), such as the Healthcare, Cost of Living, and Crime indices for certain time periods, as shown in Tables~13--16 in the Appendix. To complement this evaluation, we use Kernel Density Estimation (KDE) plots to compare the distributions of observed and imputed values, indicating that the overall distributional structure is broadly preserved. However, distributional similarity does not guarantee accurate recovery of individual observations or cross-country patterns.}

\begin{table}[htbp]
\centering
\small
\renewcommand{\arraystretch}{1.02}
\caption{\textcolor{black}{Percentage of Missing Values by Sub-Index. Refer to Tables~13--16 in the Appendix for detailed period-wise information on missingness.}}
\label{tab:missingdata}

\begin{tabular}{llc}
\toprule
\textbf{Sub-Index} & \textbf{Attribute} & \textbf{\% Missing} \\
\midrule

\multirow{6}{*}{\textbf{Economic}} 
& GDP Growth Rate & 2.82 \\
& Inflation Rate & 21.83 \\
& GDP Per Capita & 4.80 \\
& Unemployment Rate & 41.29 \\
& Cost of Living Index & 87.58 \\
& Local Purchasing Power Index & 87.58 \\
\midrule

\multirow{6}{*}{\textbf{Institutional}} 
& Control of Corruption & 4.01 \\
& Government Effectiveness & 4.48 \\
& Political Stability & 3.31 \\
& Regulatory Quality & 4.44 \\
& Rule of Law & 2.26 \\
& Voice and Accountability & 3.13 \\
\midrule

\multirow{10}{*}{\textbf{Quality of Life}} 
& Life Expectancy at Birth & 6.83 \\
& Doctors per 10,000 & 67.91 \\
& Access to Electricity & 47.84 \\
& CO$_2$ per Capita & 57.84 \\
& Gender Development Index & 66.47 \\
& Gender Equality Index & 66.50 \\
& Gender Inequality Index & 66.50 \\
& Human Development Index & 61.83 \\
& Health Care Index & 92.44 \\
& Crime Index & 88.78 \\
\midrule

\multirow{9}{*}{\textbf{Sustainability}} 
& CO$_2$ Emissions & 44.21 \\
& Non-renewable Electricity & 45.30 \\
& Renewable Electricity Output & 45.30 \\
& Micro Air Pollution & 55.78 \\
& Greenhouse Emissions & 43.07 \\
& Renewable Energy Usage & 3.45 \\
& Trade per GDP & 74.49 \\
& Total Labor Force & 74.93 \\
& Industry per GDP & 75.48 \\
\bottomrule
\end{tabular}
\end{table}

Table \ref{tab:missingdata} shows the percentage of missing values for each variable within the different sub-indices in our dataset. The identified missing values in each column underscore the need for robust imputation techniques to ensure data completeness. Specifically, we use the Random Forest Regressor \cite{rf_3}, which accounts for the relationships between variables when filling in missing values. Additionally, we use  MICE data imputation \cite{RFR}, ensuring more reliable results for subsequent analysis.

\subsection{Index creation using PCA and Factor Analysis}
 
\subsubsection{Factor Analysis}

Factor Analysis  \cite{bartholomew2011latent} is a probabilistic modelling approach that assumes the observed variables are generated from a smaller set of latent factors plus noise, thereby explicitly decomposing variance into shared and unique components. This makes FA particularly suitable for uncovering hidden structures or constructs in the dataset, such as economic indicators. 

  Factor Analysis is a linear statistical approach used to capture and explain the variability among observed variables by grouping them into underlying, unobserved variables known as factors. This technique simplifies the observed variables into fewer latent factors, highlighting relationships and reducing dimensionality\cite{Engelhardt2013}. The purpose of factor rotation is to increase the overall interpretability by converting components into uncorrelated factors. Although there are other approaches, we implement Factor Analysis using varimax rotation to identify latent structures within each sub-index. The co-variation in the observed variables is due to the presence of one or more latent variables (factors) that exert causal inference on observed variables. 

 Factor Analysis has been used on several instances to create an index. 
 Pervaiz et al.  \cite{Factor_Analysis_1} illustrated the application of factor analysis in constructing a Social Exclusion Index in Pakistan, utilising a multidimensional deprivation score derived from several indicators related to living standards and education. Luzzi et al. \cite{Factor_Analysis_2} highlighted the importance of Factor Analysis in constructing poverty indicators, showing that it can uncover insights into multidimensional poverty by merging multiple underlying variables into common factors that represent key dimensions of poverty.  Alkire and Santos \cite{Factor_Analysis_3} developed the Multidimensional Poverty Index using a dual-cutoff approach to recognise and assess poverty across various dimensions. We need to first perform an adequacy test to check if we can find the factors in our dataset before we run Factor Analysis. We implemented the Kaiser-Meyer-Olkin (KMO) test, which measures the suitability of data for factor analysis. KMO estimates the proportion of variance among all the observed variables \textcolor{black}{and measure confirmed sampling adequacy}. KMO values range between 0 and 1, where a value of less than 0.5 is considered inadequate \cite{FA}.

\subsubsection{Principal Component Analysis}

 Principal Component Analysis (PCA) \cite{jolliffe2002principal} is primarily a data reduction method that seeks orthogonal linear combinations of observed variables called principal components. This maximises the total variance in the dataset. It is a deterministic transformation that does not assume any underlying generative model and treats all variance as meaningful signal.  PCA can approximate the subspace identified by FA, but it does not distinguish between signal and measurement error. \textcolor{black}{The limitation of its interpretability in domains where latent causation is of primary interest. Consequently, PCA is often preferred for preprocessing and dimensionality reduction tasks, whereas FA is more appropriate for theory-driven analysis involving latent variables.}
 
 PCA  is a statistical method commonly used for reducing dimensionality, transforming a high-dimensional dataset into a lower-dimensional version while preserving as much variance as feasible \cite {abdi2010principal}.   PCA  uses principal components that are orthogonal linear combinations of the original features, capturing maximum variance without overlapping information \cite{Principle_Component}.  PCA has been successfully applied in different fields to create composite indices. For example, Singh \cite{Principle_Component_1} used PCA to create India's food consumption index, emphasising its ability to manage multicollinearity among variables—a frequent challenge in complex surveys. Choi et al.  \cite{Principle_Component_2}   used PCA in creating an Aggregate Air Quality Index, offering a comparative assessment of air quality across various modes of transport.  Ali et al. \cite{Principle_Component_3} used PCA for a water quality index, highlighting the method’s strength in managing missing data and outliers.

\subsection{Sub-indices}

We need to create the sub-indices using available data and indices to create the Global EoLI, where we review the quality of life over time for selected countries. \textcolor{black}{Therefore, to ensure indicators were standardised before dimensionality reduction to ensure comparability across scales, four sub-indices were created as follows.}

\subsubsection{Economic Sub-Index}

 In the Economic sub-index, we include the \textit{cost of living} since it provides an accurate representation of economic well-being \cite{Ogura}. The GDP growth rate is an important macroeconomic variable that affects stock exchange performance, underscoring its importance in evaluating the economic health of a country \cite{Verma}. There is a strong relationship between GDP per capita growth and financial growth, and quality of life  \cite{Shapoval}. We include factors such as \textit{inflation rate} and "unemployment rate" because of the trade-off between unemployment and inflation rates, indicating that changes in these economic indicators can significantly impact individuals' well-being \cite{Blanchflower}. 

\subsubsection{Quality of Life Sub-Index}

The  Quality of Life sub-index covers dimensions such as daily activities and physiological aspects, which are vital for evaluating overall well-being \cite{Behzadifar}. This indicator contains 7 sub-indicators, such as the "healthcare accessibility", which is directly related to the general well-being and enables us to observe health inequalities within and between countries, emphasising the importance of addressing disparities in healthcare access and quality \cite{Ona}. We include the  \textit{crime rate} since it affects quality of life, and policymakers and law enforcement officials can have more effective strategies for crime prevention and response \cite{Sakip}. We include the average years of schooling, in conjunction with other factors such as expected years of schooling, and life expectancy, which impact the overall quality of life  \cite{Jalil}. We also include factors such as the \textit{poverty rate} to cover the impact of financial inclusion on quality of life  \cite{Ullah}.

\subsubsection{Institution Sub-Index}

Our  Institutional sub-index comprises the \textit{corruption perception index}, \textit{freedom of speech index}, \textit{fundamental rights}, \textit{order and security}, \textit{regulatory enforcement}, \textit{civil justice}, and \textit{criminal justice}. These factors are important as they tell us about the relationship between the institutions and income inequality, which ultimately affects the ease of living in a country \cite{Behnezhad}. Factors such as the  HDI \cite{sagar1998human} measure development by considering factors such as living standards, education and sustainable development. It is important in assessing a country's progress in improving the quality of life of its population \cite{Jones}.  

\subsubsection{Sustainability Sub-Index}

Our Sustainability sub-index incorporates key indicators such as \textit{carbon dioxide emissions}, \textit{electricity production from renewable sources},  \textit{micro air pollution}, and \textit{greenhouse gas emissions}, which are critical for evaluating environmental sustainability and guiding policy decisions.  The carbon dioxide emission is a fundamental metric in assessing environmental impact, as it is directly linked to industrial activities and economic growth. Studies have shown that economic expansion often correlates with increased carbon dioxide emissions, necessitating their inclusion in sustainability assessments to accurately reflect ecological consequences \cite{Sustainability_Index_1}. In addition to carbon dioxide emissions, the percentage of electricity generated from renewable sources is vital for both developed and developing nations. Transitioning to renewable energy is essential for achieving Sustainable Development Goals (SDGs) and ensuring long-term energy sustainability\cite{Sustainability_Index_2}. Moreover, micro-level air pollution has been increasingly recognised for its detrimental effects on public health, making it a crucial factor in sustainability assessments. Studies have demonstrated that air pollution directly impacts health outcomes, underscoring the necessity of including this metric in sustainability indices \cite{Sustainability_Index_3}. The integration of air quality indicators alongside carbon dioxide emission and renewable energy production provides a holistic view of sustainability efforts, allowing for more informed decision-making and resource allocation\cite{Sustainability_Index_4}.

\subsection{Framework of Global Ease of Living Index}

Our  Global-EoLI aims to address the limitations of existing related indices (such as the Ease of Living index) by providing a more holistic view of living conditions by including factors such as healthcare access, crime rate, and freedom of speech and implementing them on a global scale. Therefore,  we offer policymakers a practical tool to identify and address specific areas needing improvement. Our contribution significantly enhances the literature on the quality of life assessment, providing a robust, reproducible methodology that can be applied across different contexts to inform targeted interventions.  We designed our index to be openly accessible, with all data and methodologies available for further research purposes, ensuring transparency and reproducibility. Moreover, our approach avoids specific biases towards any single economic factor, offering a balanced assessment that can guide policymakers in implementing well-rounded improvements in living standards. This perspective is essential for capturing the true complexity of well-being and fostering informed, effective policy decisions.

Our Global-EoLI consists of four major sub-indices, and to reflect the varying levels of importance of each category in assessing living standards, we apply different weights to each sub-index.   
The framework (Figure \ref{fig:framework}) for creating the Global EoLI is centred around the collection of various socio-economic and infrastructural data points, pre-processing them for consistency, and applying advanced statistical techniques to generate a reliable index.


\begin{figure*}[!t]
\centering
\includegraphics[width=\textwidth,keepaspectratio]{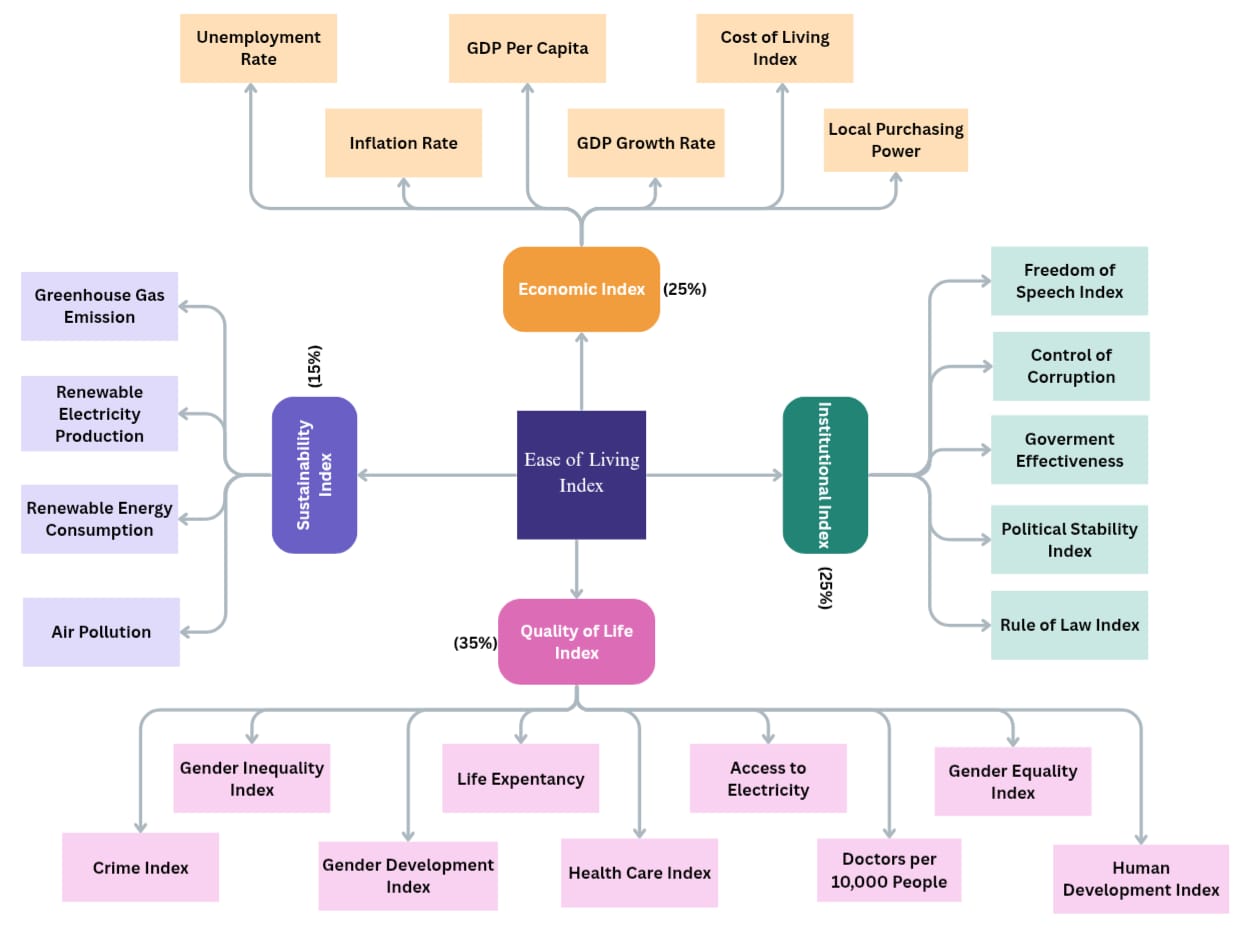}
\caption{Ease of Living Index Creation}
\label{fig:ease_of_living_index}
\end{figure*}

\begin{figure*}[!t]
\centering
\includegraphics[width=\textwidth]{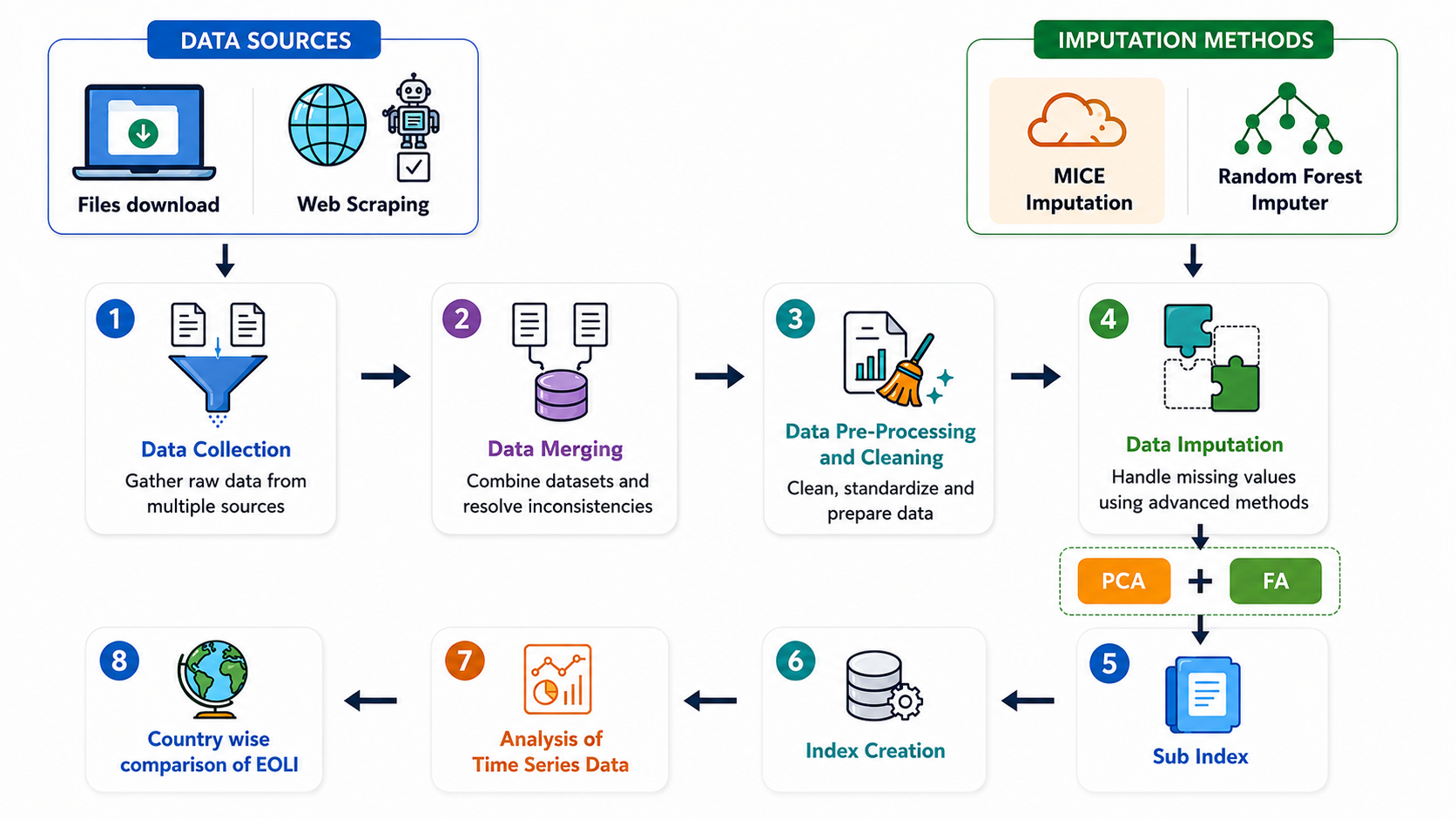}
\caption{Global Ease of Living Index Framework showing the major stages, including machine learning components such as data imputation and index creation.}
\label{fig:framework}
\end{figure*}

In Step 1 (Figure \ref{fig:framework}),  the data collection process involved both web scraping and direct downloads from reputable sources. We utilised web scraping to extract data from several websites, with BeautifulSoup \footnote{\url{https://pypi.org/project/beautifulsoup4/}} as the primary library for this task \cite{webscraping}. \textcolor{black}{We provide a detailed list of all data source websites used in this study in Appendix \ref{tab:data_sources}. Additionally, we provide selected datasets from the respective websites in the GitHub repository for this study.}

In Step 2, we merge the data from various datasets to create a single dataset having data from all 4 sub-indices. We note that the values of the indices within the data vary in different ranges with their units. 
\textcolor{black}{
In Step 3, we pre-process and standardise the original data, ensuring imputation models operate on interpretable, scale-free, meaningful values. For example, we measure GDP (per capita) in USD (nominal), life expectancy in years, and inflation as a percentage. This ordering is deliberate by preserving the natural units of each variable during imputation, and the models can learn meaningful relationships between predictors and accurately reconstruct missing observations.  We grouped the data by country and year to achieve uniformity across the sources and for strong cross-national comparisons.   
}

\textcolor{black}{In Step 4, we apply data imputation to fill the missing values as indicated in Table 1, with further details given in the Appendix (Tables 13-16). Before applying z-score standardisation, we removed all ordinal and categorical variables from the dataset, since such variables cannot be appropriately transformed using z-score standardisation. For the remaining continuous variables, we applied the standard z-score transformation, defined as $(x - \mu) / \sigma$, which rescales each variable to have a mean of 0 and a standard deviation of 1. Since we apply standardisation after imputation is complete, we compute the mean and standard deviation used in this transformation on the full, complete dataset, ensuring that no missing data bias is introduced into the scaling parameters and that no variable with a larger magnitude disproportionately influences the factor loadings. We apply the same standardisation approach before conducting PCA.}

In Step 5, after obtaining the complete data with the best imputation method, we proceed with sub-index creation, where we deploy data reduction methods such as PCA and Factor Analysis, which enable the reduction of a vast number of indicators into a manageable set of factors that best represent the ease of living across different countries. 
\textcolor{black}{
 We explicitly distinguish the roles of  PCA and Factor Analysis within the index construction framework for methodological clarity. We utilise PCA as an exploratory tool to determine the appropriate number of latent factors based on eigenvalue analysis (Kaiser criterion). It does not contribute directly to the construction of any sub-index as shown in our Framework (Figure 2). In contrast, we use Factor Analysis with varimax rotation as the primary method to extract latent dimensions and compute factor scores, which form the basis of each sub-index. This sequential application ensures that PCA informs the structure, while  Factor Analysis performs the actual dimensionality reduction and index construction, avoiding conceptual overlap between the two techniques \cite{nardo2005handbook}.
}

In Step 6, we obtain our Global EoLI, which is a weighted sum of four indices. \textcolor{black}{Specifically, we weighted the Economic sub-index as 0.25, the Institutional sub-index as 0.25, the Quality of Life sub-index as 0.35, and 0.15 for the Sustainability sub-index \cite{mohua2018ease}. We take inspiration from the Ease of Living Index created by the Government of India \cite{mohua2018ease}, which has given 0.25 to the institutional and social sub-index, 0.05 to the economic sub-index and 0.45 to the physical infrastructure sub-index.  We focus on an index having less reliance on Economic factors and focusing on the institutional and quality of life indices, which have factors such as healthcare and government effectiveness. }  In the next step, we perform a statistical and longitudinal analysis of our Ease of Living Index.

Finally, in Step 7,  we compare the countries based on their Global Ease of Living Index. We ranked the countries for each year from 1970 to 2021, with our main focus on a few of the G20 countries. We also compare countries that are currently in their developing phase, nearing the developed status and developed countries among the G20 members. This comparison between the Ease of Living Index gives a robust evaluation of the progress a country has made over the past five decades.


 \textcolor{black}{ 
Methodologically, the index follows a standard additive composite structure, where normalised indicators are aggregated using fixed weights, ensuring transparency and comparability \cite{nardo2005handbook,angelopoulos2013epi}. Factor Analysis results further support this structure, indicating that all four dimensions contribute meaningfully without any single dominant component.
We conducted a sensitivity analysis using alternative weighting schemes (including equal weights) to address potential subjectivity and include the results in  Table \ref{tab:ranking_scenarios}. Overall, this approach aligns with the multidimensional nature of development and established composite index practices \cite{north1990institutions,acemoglu2005institutions}.
}

\textcolor{black}{Although we develop the Global EoLI for the world, we show detailed analysis for selected countries. The selection of the countries was based on factors that include the population size, GDP, and their nature, which includes developing and developed states. Hence, we selected the US, Japan, Germany, the United Kingdom, Canada and Australia since they are developed countries. All the countries selected fall in the top 15 GDP by country list, except Malaysia, which represents a developing state of interest as it has a similar population to Canada and Australia for comparison. We selected India and China because they have a similar population and are developing states.}


\subsection{Technical details}

\textcolor{black}{In the case of the Random Forest imputation (\textit{missForest}), we use an ensemble of 100 decision trees to balance computational efficiency and stability of imputation as indicated by results of trial experiments. The minimum samples per split and minimum samples per leaf were set to 2 and 1, respectively, following the default configuration recommended for regression tasks, which allows trees to grow sufficiently deep to capture complex non-linear relationships in the data.  In the case of MICE imputation, we imputed each variable with missing values using a regression estimator conditioned on all other variables in an iterative chained fashion. We used Bayesian Ridge regression as the estimator for continuous variables, as it incorporates regularisation through automatic relevance determination and is robust to multicollinearity among predictors \cite{azur2011multiple}. A maximum of 10 iterations was set to ensure convergence of the imputed values across cycles, with a convergence tolerance of $1 \times 10^{-3}$, below which further updates to imputed values are considered negligible. We repeated the imputation across 5 multiple imputations to account for uncertainty in the missing data mechanism, and the mean of the imputed datasets was taken for subsequent index construction. We fixed the random state across all experiments to ensure reproducibility.}

\section{Results}

\subsection{Data imputation using MICE and RFR}

We evaluate data imputation methods to assess their ability to impute the missing values across key sub-indices. We assess their performance using the Root Mean Square Error (RMSE) and Mean Absolute Error (MAE).  We created synthetic data for evaluating the imputation model using the columns from the respective sub-indices with the lowest percentage of missing (null) values.  To simulate the missing value situation, 40\% of the values were randomly deleted from the selected columns. We subdivided this data into training and testing datasets.  We run 30 independent model training experiments with random data initialisation for each model and report the mean and standard deviation. 

Table \ref{tab:compareimp} compares  RMSE and MAE for MICE and RFR models across the Economic, Institutional, Quality of Life, and Sustainability sub-indices. In the Economic Index, MICE demonstrates a lower RMSE for GDP Growth Rate, highlighting its strength in variables with more straightforward relationships, while RFR achieves a smaller RMSE for both GDP growth rate and GDP per capita, reflecting its effectiveness in capturing more complex and straightforward data structures. Within the Institutional Index, both models exhibit similar RMSE scores across variables, with MICE performing moderately better than RFR, indicating a linear relationship between the variables. The lower values for RMSE indicate consistent and reliable imputation performance for this sub-index. In the Quality of Life sub-index, RFR shows a slight advantage in variables such as Life Expectancy at Birth and Access to Electricity, with lower RMSE scores, suggesting a better fit for imputation in cases with more diverse data distributions. In the Sustainability sub-index, RFR consistently outperforms MICE, achieving lower RMSE for all the attributes,  which underscores its capability in handling complex, non-linear environmental data. 
The MAE scores provide additional insight into model performance, particularly emphasising the absolute magnitude of errors.

\begin{table*}[htbp!]
\small
\centering
\caption{ Comparison of   MICE and RFR for given subindices, showing mean and standard deviation (in brackets) for RMSE and MAE from 30 independent model training runs.}
\label{tab:compareimp}
\resizebox{\textwidth}{!}{%
\begin{tabular}{|l|l|cc|cc|}
\hline
\textbf{Sub-Index} & \textbf{Attribute} &   \textbf{MICE} & &  \textbf{RFR} &\\
 \hline
 & & \textbf{RMSE} & \textbf{MAE} & \textbf{RMSE} & \textbf{MAE} \\
\hline
Economic Index & GDP Growth Rate & 4.51 [0.17] & 1.92 [0.03] & 3.83 [0.23] & 1.37 [0.13] \\
                              & GDP Per Capita  & 12004.46 [355.89] & 5408.87 [56.89] & 10003.01 [249.46] & 3390.69 [305.83] \\
\hline
 Quality of Life Index & Access to Electricity & 11.40 [0.44] & 3.38 [0.15] & 6.25 [0.11] & 1.84 [0.02] \\
                                     & Gender Development Index & 0.03 [0.0009] & 0.01 [0.0004] & 0.02 [0.0002] & 0.008 [0.00007] \\
                                     & Gender Inequality Index & 0.08 [0.002] & 0.03 [0.0007] & 0.05 [0.0005] & 0.017 [0.0002] \\
                                    & Human Development Index & 0.06 [0.0015] & 0.02 [0.0006] & 0.03 [0.0003] & 0.01 [0.0001] \\
                                    & LE at Birth            & 3.54 [0.12] & 1.40 [0.04] & 2.12 [0.03] & 0.85 [0.01] \\
\hline
Institutional Index & Control of Corruption & 0.32 [0.0087] & 0.13 [0.0033] & 0.49 [0.006] & 0.29 [0.0015] \\
                                & Government Effectiveness & 0.32 [0.0084] & 0.13 [0.0026] & 0.49 [0.004] & 0.29 [0.0013] \\
                                   & Political Stability    & 0.46 [0.0105] & 0.19 [0.0043] & 0.63 [0.008] & 0.39 [0.0057] \\
                               & Regulatory Quality     & 0.35 [0.0082] & 0.14 [0.0031] & 0.45 [0.007] & 0.32 [0.0026] \\
                                 & Rule of Law            & 0.30 [0.0089] & 0.12 [0.0033] & 0.47 [0.004] & 0.28 [0.0014] \\
                                 & Voice and Accountability & 0.44 [0.0070] & 0.18 [0.0031] & 0.50 [0.011] & 0.38 [0.0068] \\
\hline
Sustainability Index & Carbon Dioxide Emissions & 72.93 [3.47] & 21.65 [0.74] & 68.86 [0.58] & 15.63 [0.31] \\
                                  & Non-Renewable Electricity Production & 23.18 [0.45] & 9.91 [0.25] & 18.17 [0.17] & 8.65 [0.08] \\
                                  & Renewable Electricity Production & 4.04 [0.17] & 1.15 [0.05] & 3.08 [0.06] & 0.95 [0.01] \\
                              & Micro Air Pollution      & 10.56 [0.29] & 4.19 [0.12] & 7.57 [0.09] & 2.94 [0.03] \\
                                  & Greenhouse Emissions     & 30.84 [0.29] & 11.33 [0.12] & 24.02 [0.37] & 7.40 [0.11] \\
\hline
\end{tabular}%
}
\end{table*}

We can summarise that RFR offers a more robust and effective solution for data imputation. MICE  excels with simpler linear relationships, while RFR demonstrated superior performance in handling data imputation for complex and non-linear data structures. Therefore, we will prioritise RFR, taking advantage of its ability to capture complex relationships and achieve greater accuracy in our data imputation process. 

\begin{figure}[htbp!]
\centering
\includegraphics[width=0.5\textwidth]{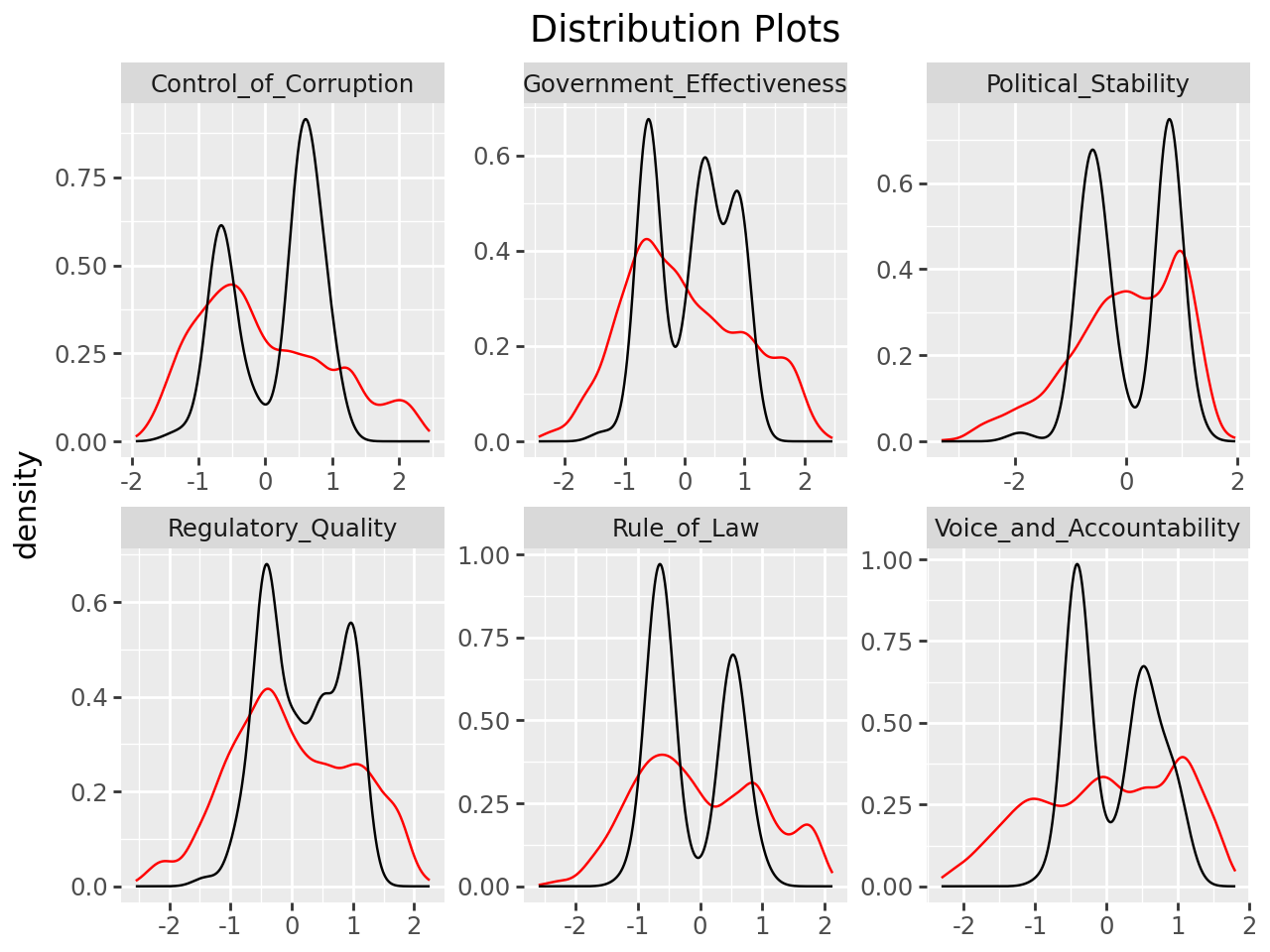}
\caption{KDE Plots for sub-indices showing governance indicators. The black curves represent the distribution of the original data, while the red curves depict the distribution of the data that has been imputed using   MICE.}
\label{fig:kde_plots}
\end{figure}

We present the  KDE  plots obtained from MICE (Figure \ref{fig:kde_plots}) to illustrate the distribution of six key governance indicators, including Control of Corruption, Government Effectiveness, Political Stability, Regulatory Quality, Rule of Law, and Voice and Accountability.  The imputed data distributions generally align well with the original distributions, indicating that the imputation process has effectively captured the underlying patterns in the data. This is further reinforced by the observation that both the imputed and original values exhibit peaks at similar points, indicating consistency in the underlying data distribution. 

Some discrepancies are evident between the original and imputed distributions, indicating areas where MICE has made adjustments to impute missing values. These differences highlight the overall structure and variability of the data, which is crucial for the reliability of the Institutional Index in subsequent analyses.

\subsection{Index creation}
 
\textcolor{black}{A key concern relates to the inclusion of variables with extremely high levels of missingness. Although imputation methods such as Random Forests and MICE are applied to reconstruct missing values, we acknowledge that variables with substantial missingness are inherently more reliant on model-based estimation than directly observed data (Table 1). To reduce the risk that such variables disproportionately influence the index, the analysis incorporates a structural safeguard through Factor Analysis. Importantly, Factor Analysis does not down-weight variables solely because they contain missing observations. Rather, after imputation and standardisation, each variable contributes to the constructed sub-index according to its empirical relationship with the underlying latent factor, as reflected by its factor loading. Indicators that show weaker shared variation with the latent construct receive lower loadings and therefore have a smaller influence on the resulting factor score. Variables with high missingness are therefore interpreted cautiously, particularly when their factor loadings are weak or unstable. Conversely, a highly missing variable may still contribute meaningfully if its imputed values remain strongly aligned with the common factor structure. Thus, construct validity is supported by examining the loading structure and KMO adequacy results, rather than assuming that missingness alone determines influence. In the Quality of Life dimension, Table \ref{tab:factor} shows that indicators such as the Human Development Index, Gender Inequality Index, and Access to Electricity have strong factor loadings, while weaker-loading indicators contribute less to the sub-index. As a result, the sub-index is primarily shaped by indicators that share a coherent latent structure, reducing the likelihood that the final index is driven by imputation artefacts.}

Table \ref{tab:pca} provides a summary of the percentage of variance explained by each principal component (PC) across the four sub-indices. In the Economic Index, PC-1 explains 44.43\% of the variance, with PC-2 and PC-3 contributing substantially. In contrast, the Institutional Index is heavily dominated by PC-1, which accounts for a significant 85.17\% of the variance. The Quality of Life Index has PC1 explaining 65.80\% of the variance, while the Sustainability Index distributes the variance more evenly across its components, with PC-1 and PC-2 together explaining over 58\%.
 
\begin{table}[htbp!]
\centering
\small
\caption{Percentage of variance explained by   PCA}
\label{tab:pca}
\begin{tabular}{|l|l|c|}
\hline
\textbf{Sub-Index} & \textbf{PC} & \textbf{\% Variance} \\
\hline
 Economic Index & PC-1 & 44.43\% \\
                                 & PC-2 & 19.35\% \\
                                 & PC-3 & 18.05\% \\
                                 & PC-4 & 12.16\% \\
                                 & PC-5 & 3.90\% \\
                                 & PC-6 & 2.11\% \\
\hline
 Institutional Index & PC-1 & 85.17\% \\
                                     & PC-2 & 6.35\% \\
                                     & PC-3 & 4.74\% \\
                                     & PC-4 & 2.02\% \\
                                     & PC-5 & 0.87\% \\
                                     & PC-6 & 0.85\% \\
\hline
 Quality of Life Index & PC-1 & 65.80\% \\
                                       & PC-2 & 10.85\% \\
                                       & PC-3 & 6.94\% \\
                                       & PC-4 & 5.44\% \\
                                       & PC-5 & 4.38\% \\
                                       & PC-6 & 3.27\% \\
\hline
 Sustainability Index & PC-1 & 37.50\% \\
                                      & PC-2 & 20.93\% \\
                                      & PC-3 & 17.76\% \\
                                      & PC-4 & 13.27\% \\
                                      & PC-5 & 10.54\% \\
\hline
\end{tabular}
\end{table}

\textcolor{black}{To interpret how the original indicators relate to the principal components, PCA loadings can be examined. Each principal component (PC) is a linear combination of the standardised original variables. For a set of indicators $X_1, X_2, \ldots, X_p$, the $j$th principal component can be expressed as:}
\begin{equation}
PC_j = w_{j1}X_1 + w_{j2}X_2 + \cdots + w_{jp}X_p
\end{equation}

where $w_{ji}$ represents the coefficient or loading associated with indicator $X_i$ on component $PC_j$. The magnitude of a loading indicates the strength of the relationship between an original indicator and a principal component, while the sign indicates the direction of association.  Table \ref{tab:pca} reports the amount of variance explained by each PC.  

\textcolor{black}{Based on the PCA variance structure, we used Factor Analysis to extract interpretable latent factors for each sub-index. PCA was used to assess the distribution of explained variance and support the dimensionality of the data, while Factor Analysis was used to estimate the factor loadings and construct the sub-index scores.}  

 

\begin{table*}[htbp!]
\centering  
\small

\begin{tabular}{l|l|c|c|c|c|}
\hline
\textbf{Sub-index} & \textbf{Indicators} & \textbf{FL-1} & \textbf{FL-2} & \textbf{FL-3} & \textbf{FL-4} \\
\hline
Eco. &  GDP Growth Rate & -0.043 & 0.037 & \fbox{0.726} & -0.131 \\
&  Inflation Rate & -0.086 & 0.144 & 0.008 & 0.043 \\
&  GDP Per Capita & 0.519 & -0.275 & \fbox{0.74} & 0.042 \\
& Unemployment Rate & -0.036 & -0.111 & -0.221 & \fbox{0.095} \\
&  Cost of Living Index & 0.541 & -0.238 & \fbox{0.681} & 0.055 \\
&   Local Purchasing Power & 0.567 & 
-0.388 & \fbox{0.586} & -0.009 \\
\hline
Ins. &  Control of Corruption & \fbox{0.912} & -0.249 & 0.186 & 0.058 \\
& Government Effectiveness & \fbox{0.869} & -0.401 & 0.203 & -0.048 \\
& Political Stability & \fbox{0.811} & -0.206 & 0.089 & 0.237 \\
& Regulatory Quality & \fbox{0.857} & -0.378 & 0.183 & -0.07 \\
& Rule of Law & \fbox{0.924} & -0.303 & 0.158 & 0.039 \\
 & Voice and Accountability & \fbox{0.842} & -0.231 & 0.032 & 0.192 \\
 \hline
& Life Expectancy & 0.363 & 0.234 & 0.116 & -0.025 \\
Life& Doctors Per 10,000 &0.346 &  \fbox{0.412} & 0.205 & 0.067 \\
& Access to Elec.&  \fbox{0.286} &  -0.857 & 0.007 & -0.006 \\
& CO2 per capita  & 0.299 & -0.428 & 0.311 & -0.105 \\
& Gender Development & 0.226 & \fbox{0.49} & -0.01 & 0.206 \\
& Gender Inequality Index & -0.467 & \fbox{0.716} & -0.241 & -0.05 \\
& HDI & 0.443 & \fbox{0.866} & 0.172 & 0.037 \\
& Health Care & 0.31 & \fbox{0.367} & 0.117 & 0.013 \\
& Crime Index & -0.375 & \fbox{0.65} & -0.173 & -0.03 \\
\hline 
Sus. & CO2 emissions & 0.025 & 0.043 & 0.11 & \fbox{0.667} \\
& Electricity: non-renewable & -0.071 & -0.136 & -0.043 & \fbox{0.46} \\
& Electricity: renewable & 0.267 & -0.049 & 0.106 & \fbox{0.533} \\
& Micro air pollution & -0.328 &  \fbox{0.164} & 0.007 &-0.549 \\
& Greenhouse emission & -0.079 & 0.009 & 0.055 & \fbox{0.757} \\
\hline
\end{tabular}

\caption{Factor Loadings for indicators in sub-index using Factor Analysis. \textcolor{black}{We highlighted the leading values and presented indicators for the subindices: Economic (Eco.), Institutional (Ins.), Quality of Life (Life), and Sustainability (Sus.).}}
\label{tab:factor}
\end{table*}
\textcolor{black}{The factor loadings shown in Table\ref{tab:factor} are not used directly as country rankings. Instead, they indicate how strongly each standardised indicator is associated with an underlying latent factor. After Factor Analysis, we compute factor scores for each country-year observation using the standardised indicator values and factor-score coefficients derived from the factor model. Indicators with stronger loadings contribute more strongly to the corresponding latent dimension, while negative loadings indicate inverse relationships with that factor.}

\textcolor{black}{As shown in Table 4, FL-1 is interpreted as the Institutional factor, FL-2 as the Quality of Life factor, FL-3 as the Economic factor, and FL-4 as the Sustainability factor, based on the dominant loading patterns. The numbering of factors does not represent their ranking or importance. For example, FL-4 does not mean that Sustainability is the least important factor; it only indicates that Sustainability is represented by the fourth extracted factor. The sub-index scores derived from these factors are then used to rank countries within each dimension. We compute the final Global EoLI by aggregating the four sub-index scores using the predefined weights assigned to the Economic, Institutional, Quality of Life, and Sustainability sub-indexes.
}

As shown in Table \ref{tab:factor},   Factor 1 includes high loadings on indicators such as  Control of Corruption, Government Effectiveness, and Rule of Law, encapsulating governance quality, suggesting a strong association with the institutional index. Factor 2 encompasses social and human development aspects, as observed by significant loadings on the Gender Inequality Index, Human Development Index, and Access to Electricity. This factor highlights areas related to social equity and basic infrastructure, which corresponds most with our Quality of Life Index. Factor 3 captures economic prosperity, with high loadings on GDP Growth Rate, GDP Per Capita, and Cost of Living Index, emphasising indicators tied to wealth and living standards, which are similar to our Economic Index. Lastly, Factor 4 reflects environmental sustainability, focusing on emissions, renewable energy, and pollution, with high loadings on Carbon Dioxide Emissions and Greenhouse Emissions, which correspond to the Sustainability Index. We use these factors to build 4 sub-indices that make up our Global EoLI. 

 \textcolor{black}{
Factor Analysis reduces dimensionality by identifying patterns of correlation among the variables, and in our case, it generated 4 distinct factors (corresponding to the sub-indexes) that capture key dimensions of the dataset. These factors represent the underlying sub-indices, and the factor loadings help clarify their structure, reinforcing our assumption that only four factors are needed to effectively explain the data. We revisit that Factor Analysis in our framework is used to identify and interpret latent constructs for sub-index construction, while PCA is applied separately to assess overall variance explained and validate dimensionality. These methods are not interchangeable, but are used sequentially.
}
 
\textcolor{black}{We note that, as shown in Table 1, the Quality of Life and Sustainability indicators contain a relatively high proportion of missing data. In addition, Table \ref{tab:kmo} shows that the Sustainability Index has the lowest KMO value, indicating weaker sampling adequacy and a less cohesive factor structure compared with the other sub-indices. Table \ref{tab:factor} shows that the Sustainability indicators load primarily on FL-4, confirming that these indicators form a distinct environmental sustainability factor. However, FL-4 should not be interpreted as the least important factor solely because it is labelled as the fourth factor. Rather, high loadings on FL-4 indicate that these indicators strongly define the Sustainability factor. The comparatively smaller influence of Sustainability on the final EoLI ranking arises from its lower assigned sub-index weight in the final composite index and from the need to interpret this dimension cautiously due to its lower KMO value and higher missingness. Thus, Sustainability contributes as a separate dimension of the EoLI, but its structural reliability is weaker than the Institutional and Quality of Life dimensions.
}

\begin{table}[htbp!]
\centering
\small
\caption{ Results for  KMO test in  Factor Analysis by two data imputation methods}
\label{tab:kmo}
\begin{tabular}{|l|c|c|}
\hline
\textbf{Sub-Index} & \textbf{MICE} & \textbf{Random Forest} \\
\hline
Economic Index         & 0.72 & 0.71 \\
Institutional Index    & 0.90 & 0.90 \\
Quality of Life Index  & 0.89 & 0.85 \\
Sustainability Index   & 0.62 & 0.49 \\
\hline
\end{tabular}
\end{table}

 \textcolor{black}{The KMO\cite{shrestha2021factor} test assesses whether the data are adequate for Factor Analysis, with values above 0.6 generally indicating suitability and values above 0.8 indicating strong sampling adequacy. As shown in Table \ref{tab:kmo}, the Economic, Institutional, and Quality of Life sub-indices show acceptable KMO values under both MICE and RFR imputation. However, the Sustainability sub-index has a KMO value of 0.49 under RFR-imputed data, which is below the conventional adequacy threshold. This indicates that the Sustainability indicators have a weaker and less cohesive factor structure when RFR imputation is used. The MICE-imputed Sustainability data produce a KMO value of 0.62, which meets the minimum adequacy requirement. Therefore, the MICE-based factor-analytic specification is treated as the preferred specification for the Sustainability sub-index, while the RFR-based Sustainability result is interpreted cautiously and used only as a robustness comparison. This limitation is important because the Sustainability dimension contains heterogeneous environmental and economic indicators, several of which also have high missingness. Consequently, Sustainability is retained as a theoretically important dimension of the Global EoLI, but its structural reliability is weaker than that of the Institutional and Quality of Life dimensions.
}

\subsection{Data Analysis of Sub-Indices}

We use data imputation (MICE and RFR) and dimensionality reduction  (PCA and RFR) to create data for sub-indices that provide a comprehensive view of global living conditions, spanning from 1970 to 2021. It includes data for 70 countries, capturing a wide range of economic, social, and environmental factors that influence the ease of living across different regions.  We publish our sub-indices using a GitHub repository included at the end of this paper.

\begin{figure}[htbp]
    \centering
    \includegraphics[width=0.5\textwidth]{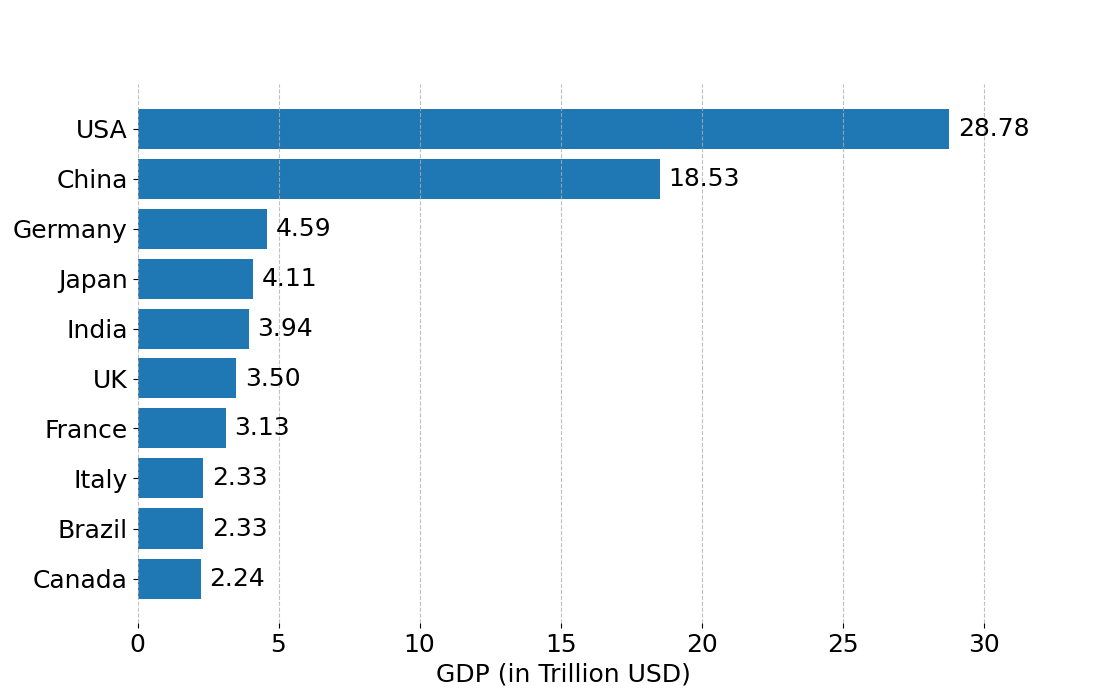}
    \caption{The plot highlights the GDP of the world's top 10 economies in 2024.}
    \label{fig:gdp_top10}
\end{figure}

We then focus on selected countries from the top 10 largest economies in the world as of 2024, which include the United States, China, Japan, Germany, India, the United Kingdom, and Canada \cite{forbes2024top}. These economies play a pivotal role in shaping global trends and offer valuable insights into how different nations approach the challenge of improving living standards amidst varying levels of institutional strength and environmental policies. This pool of countries gives us diversity in the context of the development of a nation. Countries such as the United States, Germany, Canada and Japan are considered developed countries, and countries such as India and China are considered developing nations. We have also included some of the emerging economies of the world, such as Malaysia. They are on track to become a high-income nation by the end of 2028 \cite{worldbank2021malaysia}. We have also included Australia in our analysis as it has one of the highest living standards, including the human capital index and high GDP per capita \cite{worldbank_australia_data}. 

\subsubsection{Analysis of Quality of Life sub-index}

\begin{figure*}[htbp!]
\centering
\includegraphics[width=0.75\textwidth]{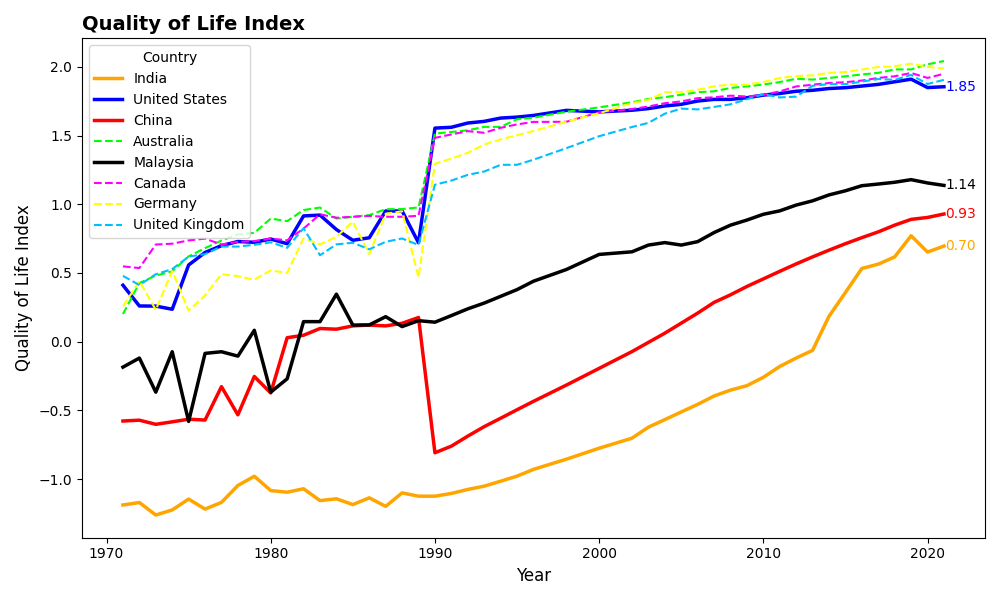}
\caption{Quality of Life sub-index over time.}
\label{fig:QOLI_over_time}
\end{figure*}

We present a longitudinal study of the Quality of Life sub-index for the selected countris. 
Figure \ref{fig:QOLI_over_time} illustrates how developed nations (the United States and Australia) have maintained consistently high living standards, while emerging economies   (India, China and Malaysia) have seen significant improvements over the years. These results reinforce that countries with better socio-economic factors, such as healthcare facilities \cite{QOLI_Results_1},   gender development \cite{QOLI_Results_2},  and HDI \cite{QOLI_Results_3} have better living conditions. On the other hand, developing countries such as China have shown a steady rise in their Quality of Life Index, reflecting improvements in infrastructure, healthcare, and education. This is all a result of the policies that the Chinese government have implemented in recent years\cite{QOLI_Results_4}. India’s Quality of Life has risen more sharply since the 1990s, as the country began to focus more on social and economic reforms\cite{QOLI_Results_5}. This analysis highlights the evolution of living standards globally and the efforts made by developing countries to catch up with established economies.


\begin{figure*}[htbp!]
\centering
\includegraphics[width=0.95\textwidth]{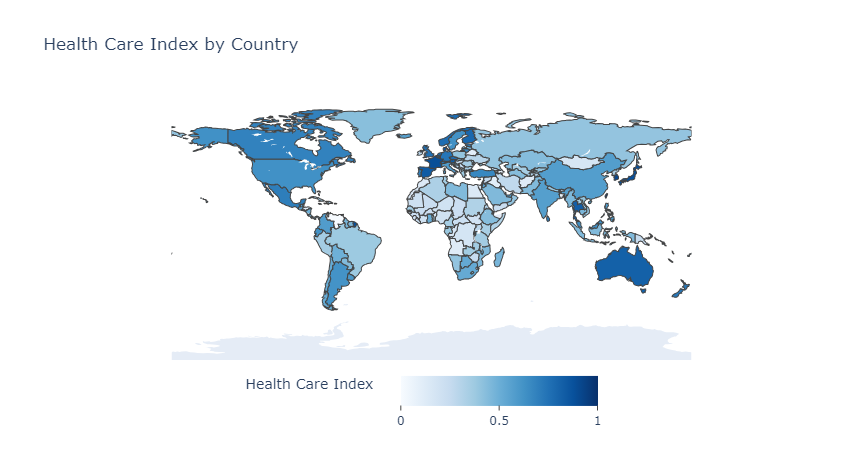}
\caption{Healthcare Index by Country (2022). }
\label{fig:healthcare_index}
\end{figure*}

Figure \ref{fig:healthcare_index} represents the choropleth map, which offers a geographical perspective on healthcare quality across different countries. The darker shades on the map represent nations with stronger healthcare systems, such as Australia, Canada, and many parts of Europe \cite{barber2017haq}, while lighter shades reflect countries with weaker healthcare infrastructures, particularly in regions such as  Sub-Saharan Africa and parts of Asia. The map visually highlights global disparities in healthcare access and quality, emphasising how economically stronger countries can provide better healthcare services and the practices \cite{joumard2010health} and policies that these countries have implemented can become a guideline for developing nations to improve their healthcare infrastructure \cite{QOLI_Results_6}. It also underscores the importance of healthcare in determining a nation’s quality of life, as countries with higher healthcare scores generally exhibit better living standards overall.
\smallskip


\subsubsection{Economic sub-index}

\begin{figure}[htbp!]
\centering
\includegraphics[width=\columnwidth]{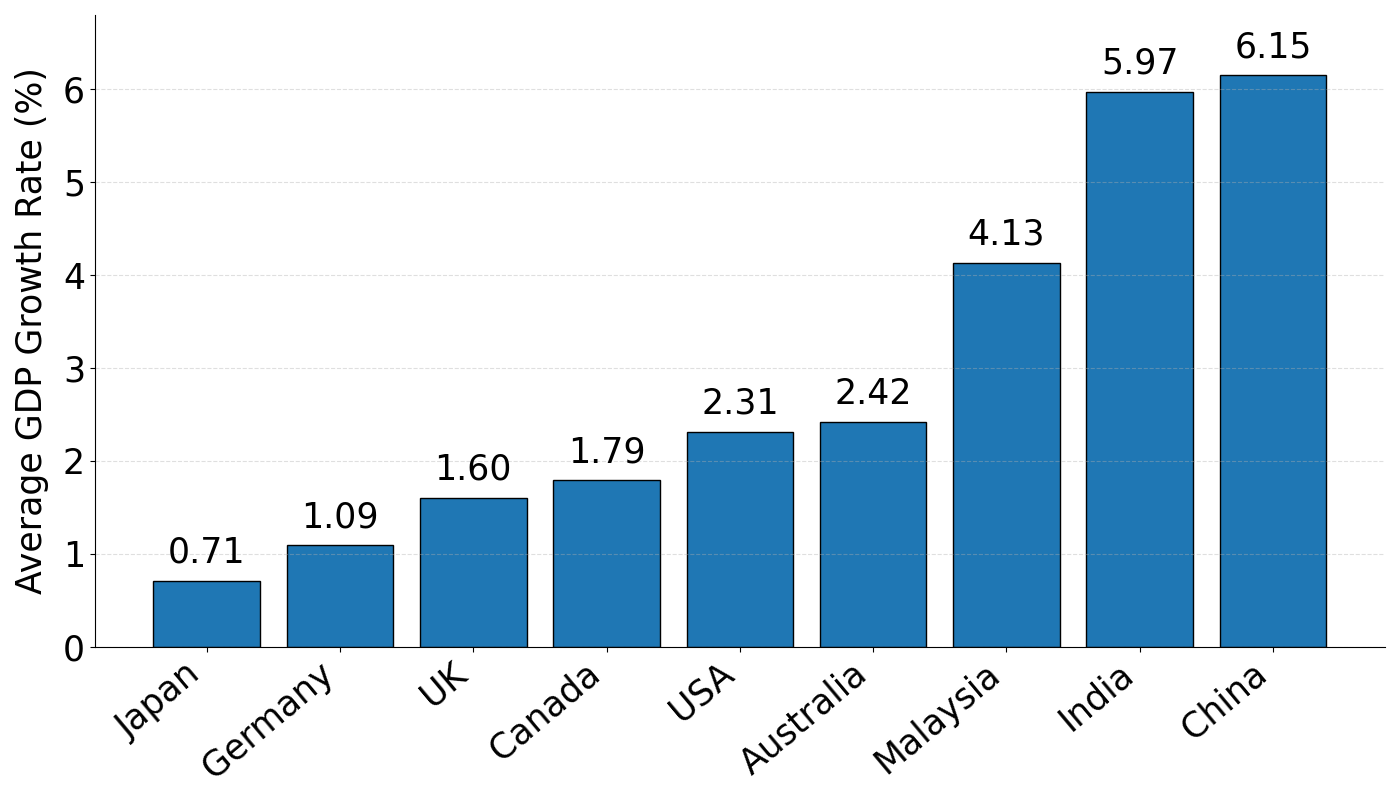}
\caption{Average GDP growth rate over the last decade.}
\label{fig:GDP_growth}
\end{figure}

Figure \ref{fig:GDP_growth} depicts the GDP growth rate graph, which highlights the average economic growth across countries in our study. We can observe that China leads the group, with an average growth rate of 6.11\%, followed by India at 5.73\%, which indicates that emerging economies have been experiencing rapid economic expansion. This is contrary to established economies \cite{Economic_results_1} such as  Germany  (1.12\%) and Japan (0.77). Countries such as the United States and Australia show moderate growth rates, reflecting their mature but stable economies. The differences in growth rates reflect each country’s stage of economic development, with emerging economies growing faster due to industrialisation and modernisation \cite{Economic_results_2}, while developed countries had slower and steadier growth \cite{Economic_results_3}.

\begin{figure}[htbp!]
\centering
\includegraphics[width=0.45\textwidth]{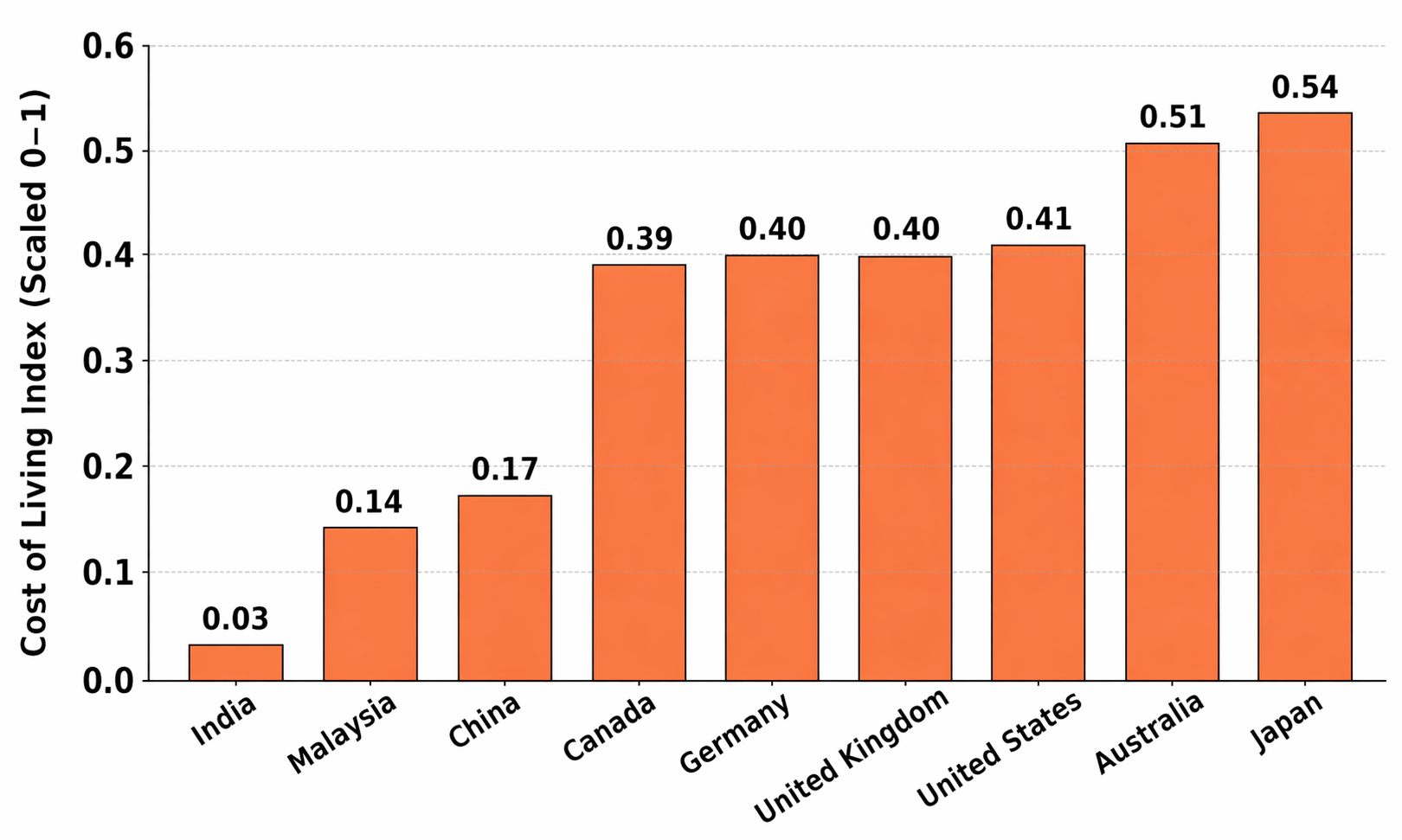}
\caption{ The 2021 Cost of Living sub-index for selected countries, illustrating variations in living expenses across different regions.}
\label{fig:cost_of_living}
\end{figure}

Figure \ref{fig:cost_of_living} represents the Cost of Living sub-index for the year 2021, which compares the relative living costs across different countries. We can observe that the United States (0.581) and Australia (0.58) have the highest cost of living, which aligns with their high GDP per capita and other advanced economies. In contrast, India (0.039), has the lowest cost of living, followed by Malaysia. In the case of India, the ratio of GDP per capita to the cost of living is the highest among all nations, which suggests that even though absolute earnings for India are much lower,  the purchasing power remains relatively high \cite{Economic_results_4}. The lower cost of living in India reflects more affordable goods, services, and housing. It also indicates that individuals can achieve a relatively comfortable standard of living, with lower income levels when compared to developed countries such as the United States and Australia, where higher incomes are offset by the elevated cost of living. This affordability is one of the key reasons for India's labour force and growth of foreign direct investment, especially for investors and expats  seeking a high quality of life at a lower cost \cite{Economic_results_5}. Developed economies such as  Germany, Japan, and the United Kingdom also show moderate cost-of-living levels, reflecting the balance between higher wages and more affordable services. This further highlights how economic development affects the affordability of goods and services, with wealthier nations tending to have higher living costs due to higher wages, property values, and general expenses.

\begin{figure*}[h]
\centering
\includegraphics[width=0.75\textwidth]{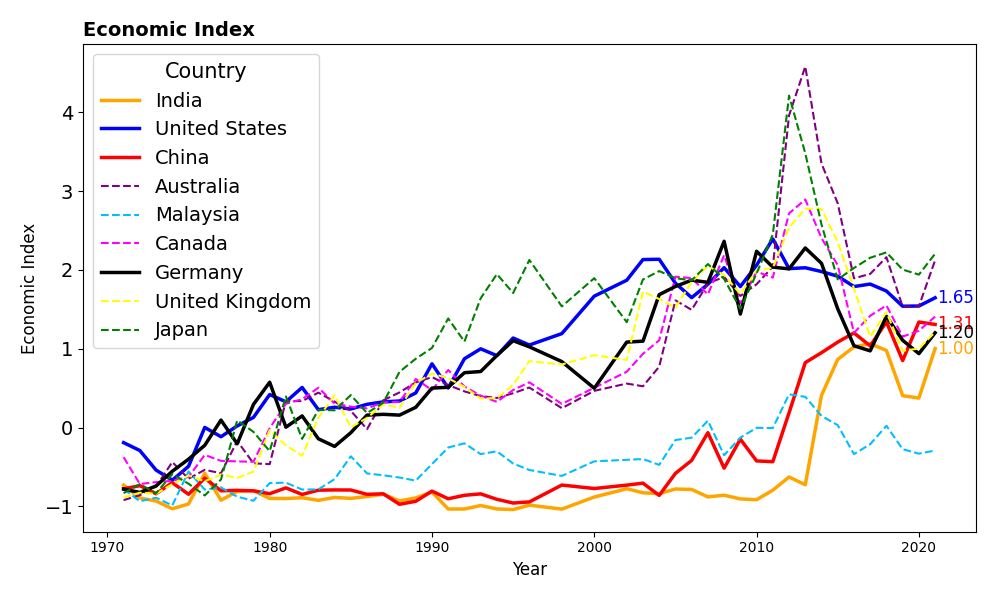}
\caption{Economic Index over the past two decades.}
\label{fig:Econ_Index}
\end{figure*}

Figure \ref{fig:Econ_Index} shows the Economic sub-index, which compares the overall economic strength of these nations over time, combining factors such as GDP, employment, inflation, and other economic indicators. China and India show rising economic stability, with China demonstrating the most significant growth, particularly around 2010, reflecting its rapid industrialisation and global influence \cite{Economic_results_6}. Australia, Canada, and Germany also show consistent performance, though at a more moderate level, reflecting their stable, well-developed economies. The United States has been one of the most consistently performing economies. India has shown improvement after 2005, signalling its growing presence on the global economic stage. This comparison highlights the dynamic nature of economic growth, where emerging economies such as India and China are catching up with the developed countries.

\subsubsection{Institutional sub-index}

The Institutional sub-index provides a comprehensive view of various governance-related factors, such as the effectiveness of rule enforcement, freedom of speech, crime rates, and other institutional structures that contribute to the overall stability and prosperity of a nation.

\begin{figure}[htbp!]
\centering
\includegraphics[width=0.5\textwidth]{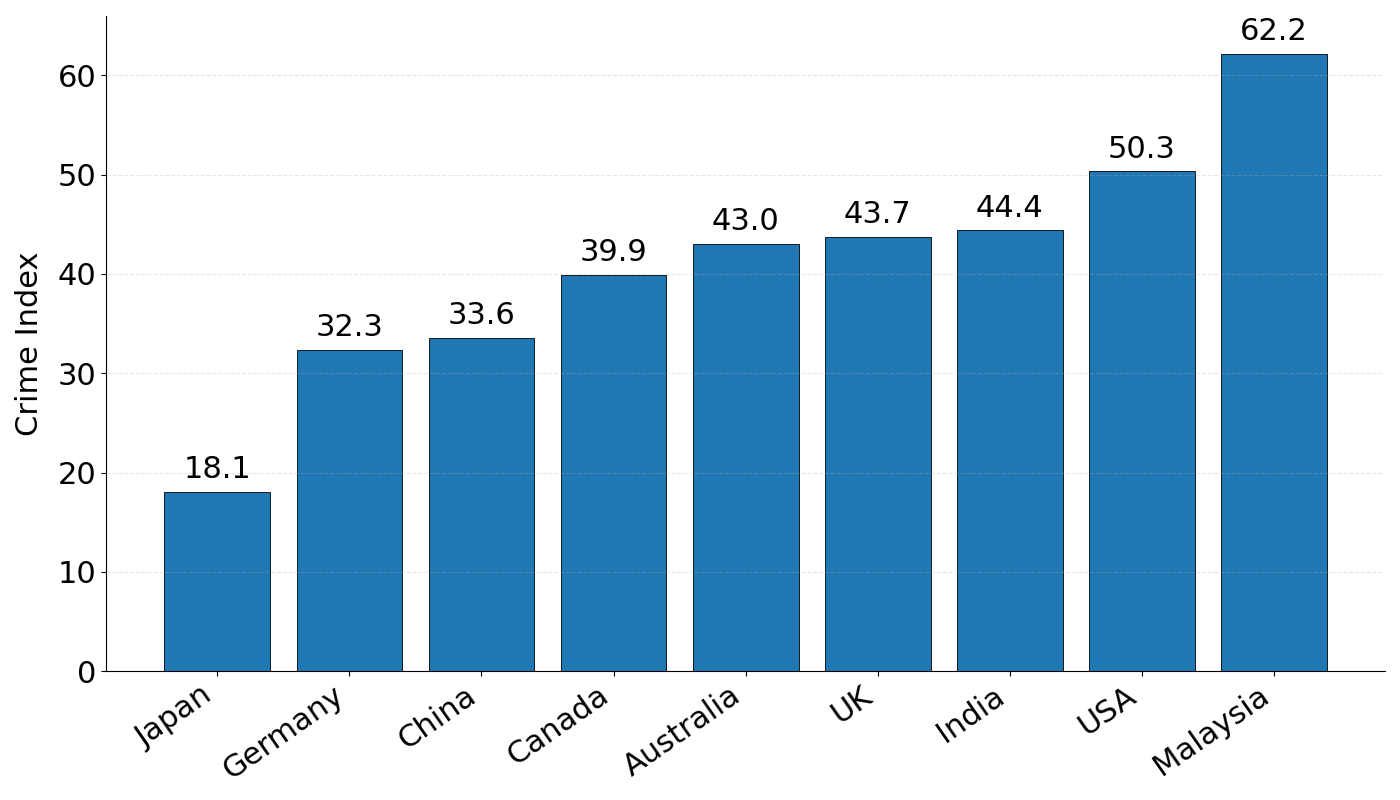}
\caption{ The crime index from 2000 to 2024 for selected countries shows Malaysia and India having higher index values compared to Japan and Germany.} 
\label{fig:crime_index}
\end{figure}

Figure \ref{fig:crime_index} presents the crime  index that provides average crime rates across the selected countries (2000-2024). We can see that Malaysia has the highest crime index \cite{Institutional_results_1}, followed by India and the United States.  The United States has the highest gun ownership in the world, with estimates suggesting there are more guns than people in the country, at around 120.5 guns per 100 residents \cite{Institutional_results_2}. The high availability of firearms has been strongly linked to the elevated rates of gun-related crimes, such as homicides, assaults, and mass shootings. China and the United Kingdom follow closely, showing relatively high crime levels but still lower than  Malaysia, India, and the United States.

 Australia, Canada, and the United Kingdom fall into the mid-level category, which likely benefits from better institutional frameworks and governance but still face challenges in fully minimising crime rates \cite{Institutional_results_3}. On the lower end, Germany and Japan exhibit the lowest crime indices among the selected countries. These nations are known for their strong law enforcement, efficient legal systems, and social stability \cite{Institutional_results_4}.

\begin{figure*}[h]
\centering
\includegraphics[width=0.75\textwidth]{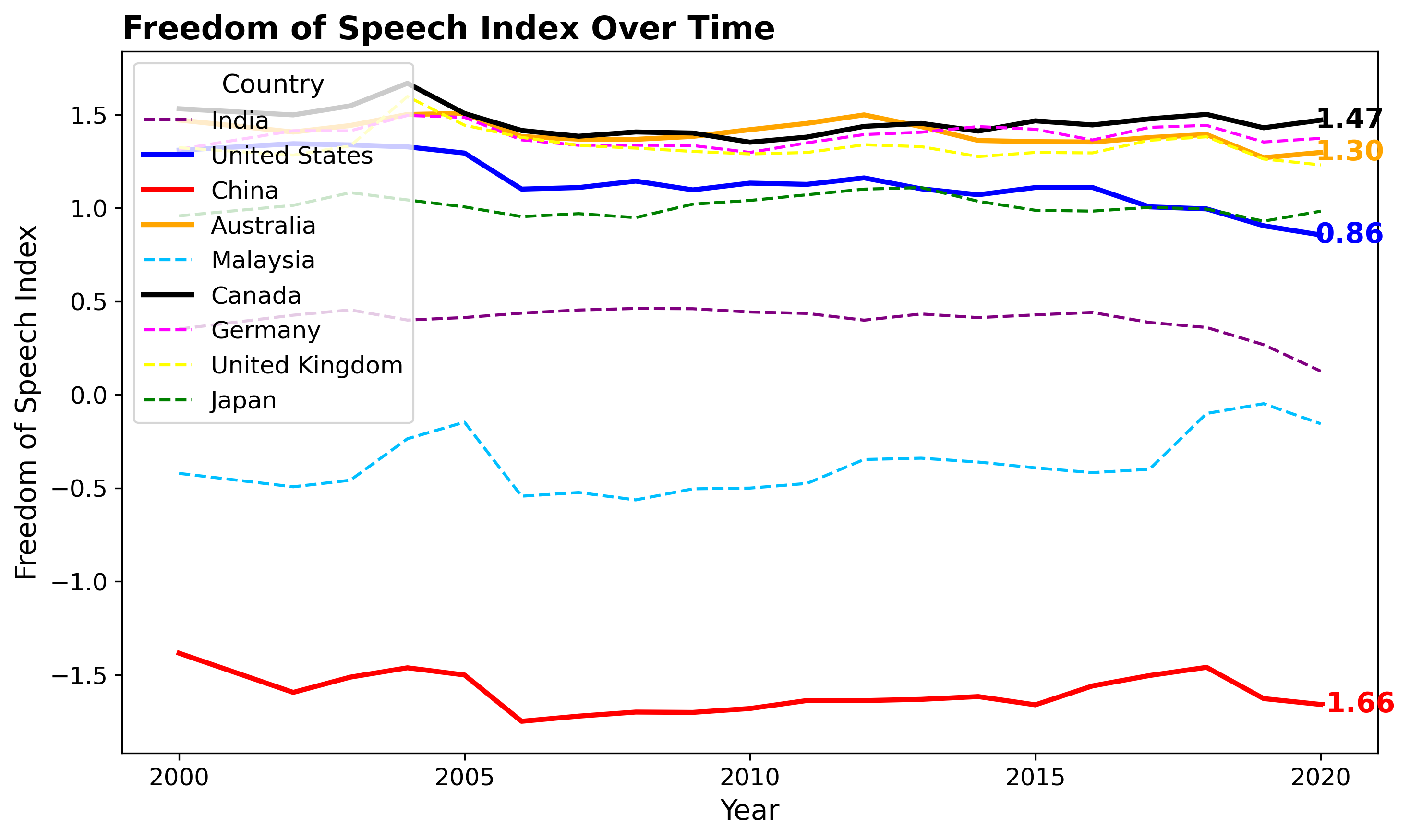}
\caption{  Freedom of Speech Index   (2000 to 2020) illustrating China's consistently lower values when compared to countries like the United States and Australia. }
\label{fig:speech}
\end{figure*}

Figure \ref{fig:speech} shows the Freedom of Speech Index, which measures the extent to which citizens in different countries can express themselves freely without fear of retribution. We observe a clear distinction between  Australia, Canada, and Germany, where the index remains consistently high, reflecting strong protections for freedom of speech \cite{Institutional_results_5, Institutional_results_6}. On the other hand, China and Malaysia consistently show lower values, highlighting restricted freedom of expression. China's index hovers around -1.5, which aligns with its authoritarian governance structure \cite{Institutional_results_7}. India occupies a middle ground, showing a stable but lower freedom of speech index than the developed nations. 

\begin{figure*}[htbp!]
\centering
\includegraphics[width=0.75\textwidth]{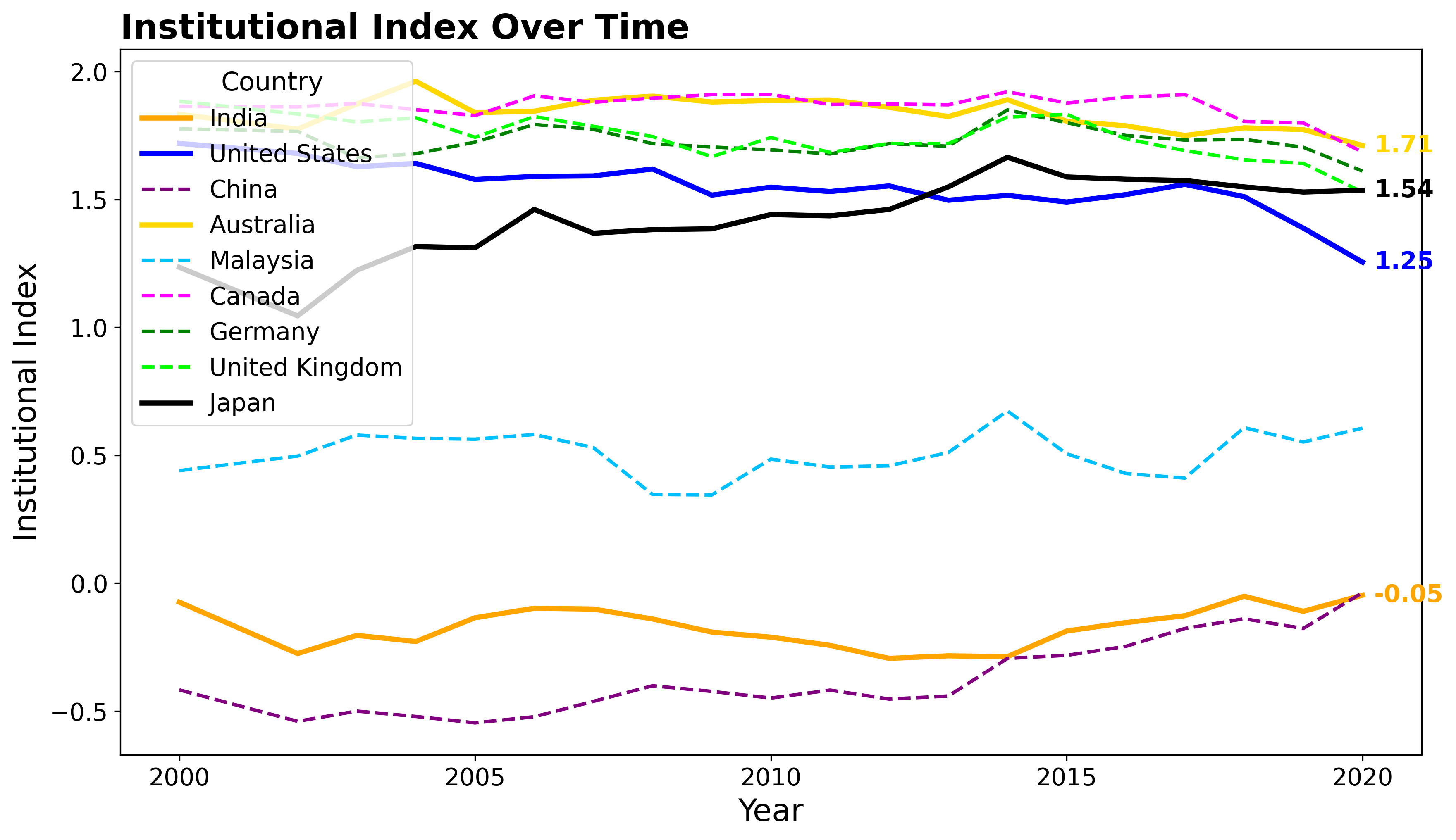}
\caption{ Institutional Index trends for selected countries (2000 to 2020), highlighting stable high performance in  Japan and gradual improvements in India.}
\label{fig:inst}
\end{figure*}

 Figure \ref{fig:inst}  shows that   Australia, Canada, and the United States have consistently maintained high institutional index scores reflecting stable governance, strict rule of law, and effective administration. India and Malaysia, on the other hand, show moderate but improving scores over time, indicating progress in strengthening governance structures. China presents a relatively lower Institutional Index, which reflects its centralised governance system and weaker rule of law. Notably, Germany and the United Kingdom remain at the top of the Institutional Index, demonstrating strong, consistent governance and regulatory frameworks. This overall analysis shows that stronger institutions often correlate with higher economic and social stability, while weaker institutions might struggle to enforce laws, protect freedoms, and maintain low crime rates.

\subsubsection{Sustainability Index}

\begin{figure*}[!t]
\centering

\includegraphics[width=14cm,height=8cm]{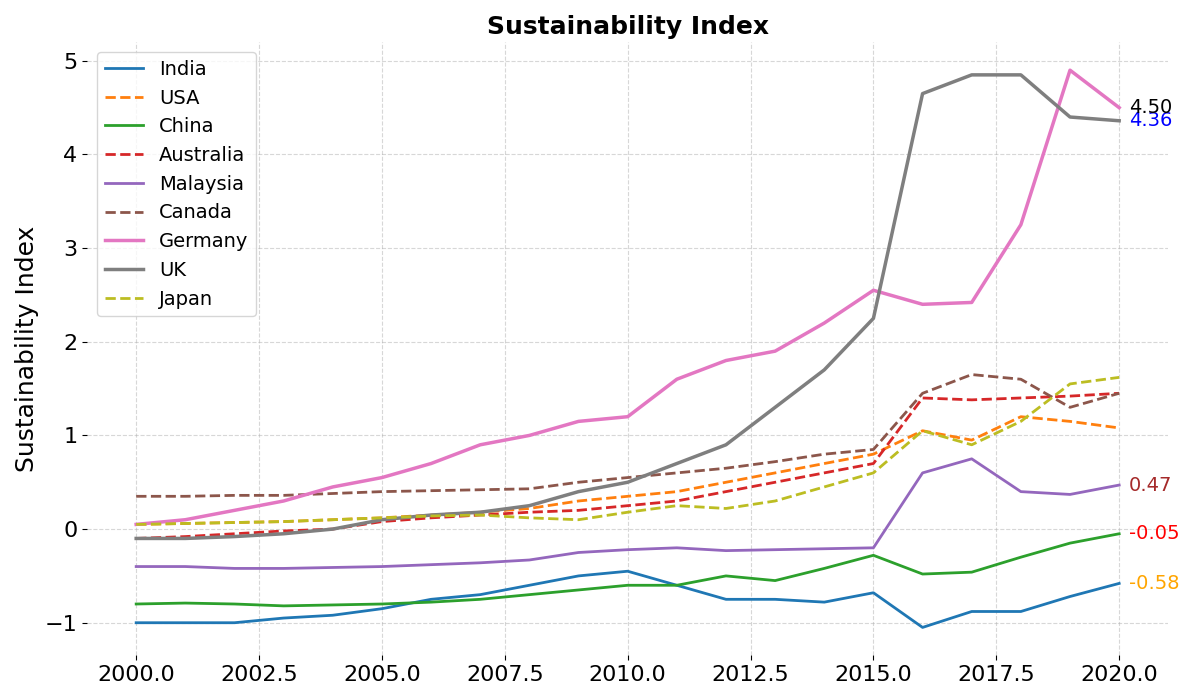}
\caption{Sustainability Index over the last two decades.}
\label{fig:sustainability_index}

\includegraphics[width=11cm,height=7cm]{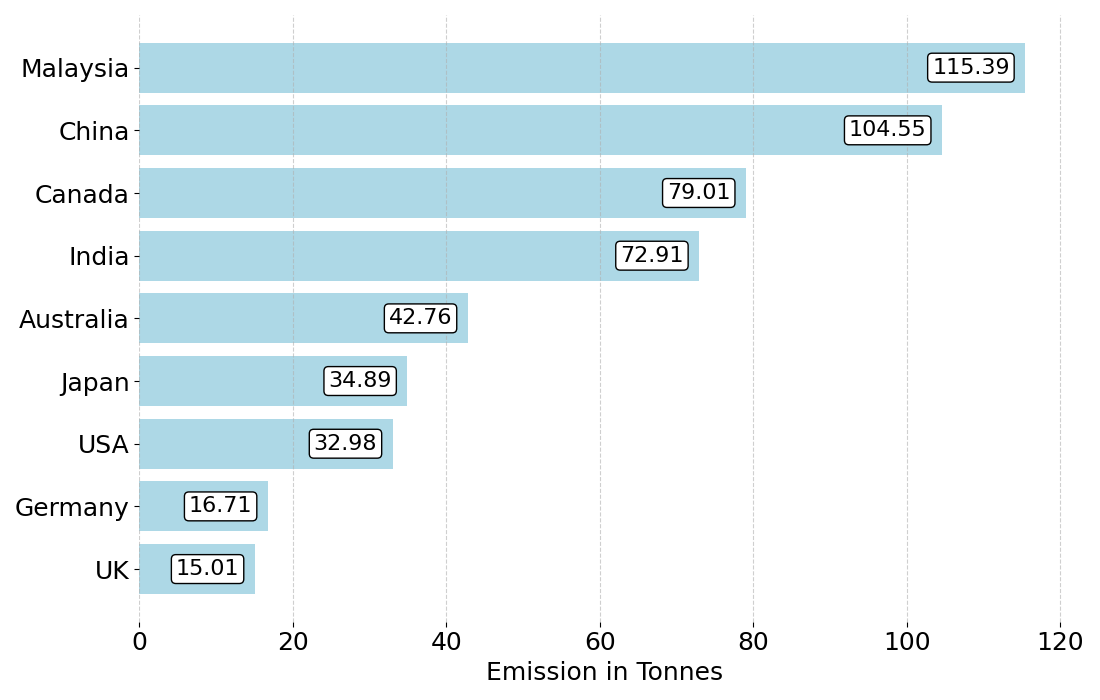}
\captionof{figure}{Greenhouse gas emissions (in metric tonnes). Malaysia and China lead in emissions when compared to lower values observed in Germany and the United Kingdom.}
\label{fig:green_house}

\end{figure*}

Figure \ref{fig:sustainability_index} presents the sustainability index \cite{} for the last two decades, which highlights a general upward trend for most countries, indicating progressive improvements in sustainable practices over the years. Developed nations such as Germany and the United Kingdom show significant gains, particularly from 2010 onwards \cite{Sustainability_results_1}, which can be attributed to rigorous environmental policies, widespread investment in renewable energy, and a societal shift towards sustainable living practices. For example, Germany's Energiewende (energy transition) initiative 
has accelerated the adoption of renewables, while the United Kingdom has committed to reducing carbon emissions through stringent regulations and clean energy incentives \cite{Sustainability_results_2}. 

On the contrary,  India and China have shown slower progress, largely because they relied on industrial growth fuelled by non-renewable energy sources and slower policy adoption in sustainability. The lower sustainability index for India can be attributed to its overreliance on coal as the primary energy source, accounting for approximately 71\% of electricity generation, further complicating the transition to a more sustainable energy model \cite{Sustainability_results_3}. The rapid rise observed in Malaysia around 2015 likely reflects targeted government programs to boost environmental standards and embrace renewable energy, possibly driven by international pressure and economic incentives \cite{Sustainability_results_4}. Overall, this index underscores the varying levels of commitment to sustainability across nations, with developed countries setting the pace and emerging economies making steady but slower advancements.

Figure \ref{fig:green_house} presents carbon emissions by selected countries, which reveals a critical challenge for many countries, such as restricting pollution while striving for economic growth.  China and India have emission levels that align with their heavy reliance on fossil fuels for industrial output and energy production \cite{Sustainability_results_5}.

In contrast, the United Kingdom and Germany have successfully implemented measures to curb emissions, due to their transition towards renewable energy and policies for reducing their carbon footprint. These policies include the decommissioning of coal-fired power facilities, more stringent automobile laws, and emissions trading programs. The importance of energy and environmental policy is shown by the inverse connection between emissions and sustainability initiatives. To improve their Sustainability Index rankings, high-emission countries must invest in cleaner technologies and adopt more aggressive environmental regulations\cite{Sustainability_results_6}.

\subsection{Global Ease of Living Index}

\begin{figure}[htbp!]
\centering
\includegraphics[width=0.5\textwidth]{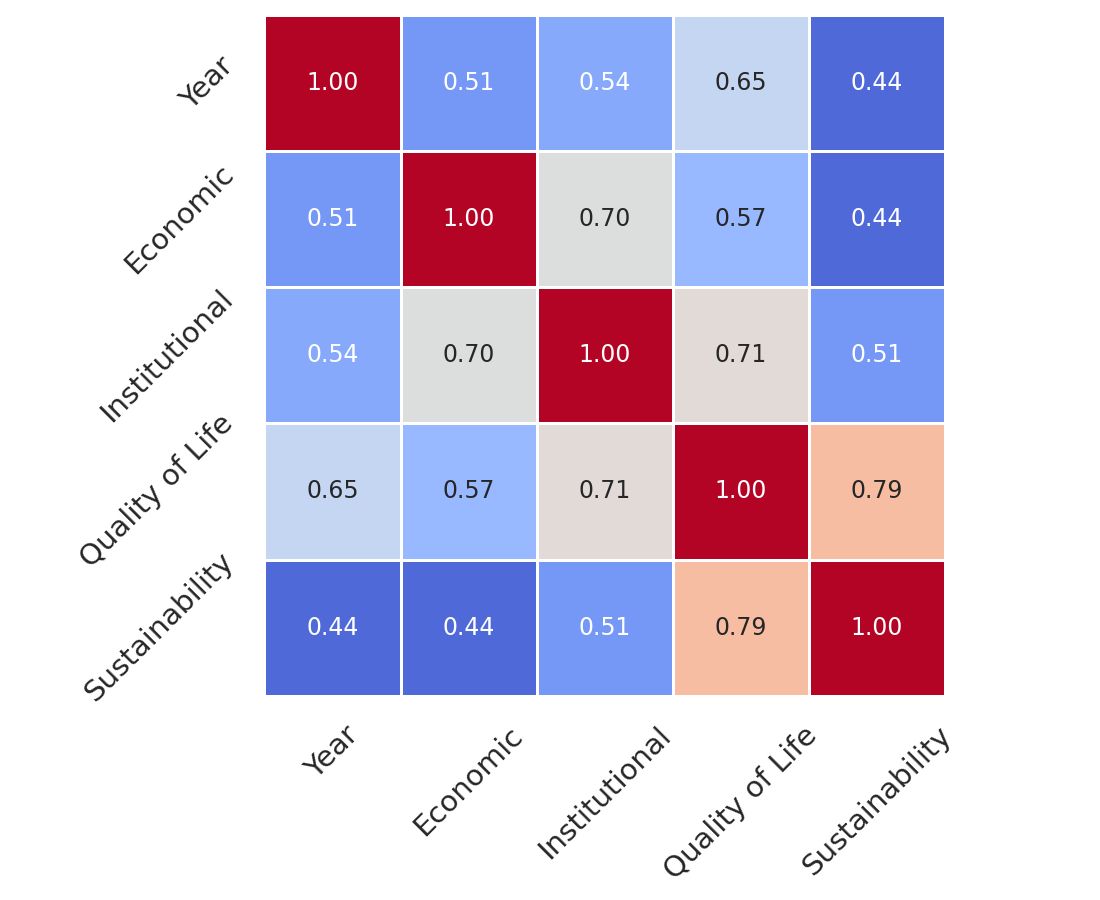}
\caption{The correlation among sub-indices used for the Global EoLI, highlighting strong correlations between the Year, Economic, Institutional, Quality of Life, and Sustainability sub-indices.}
\label{fig:heatmap_EOLI}
\end{figure}

We finally review the four sub-indices obtained from previous investigations and then develop the Global Ease of Living Index.
Figure \ref{fig:heatmap_EOLI} presents a correlation matrix of all the sub-indices, highlighting the relationships between the Economic Index, Institutional Index, Quality of Life Index, and Sustainability Index. The correlation values range between -1 and 1, where values close to 1 indicate a strong positive correlation, while values near 0 suggest little to no relationship \cite{stewart1968general}. 
A notable finding is the strong correlation between the Quality of Life and the Sustainability sub-indices, suggesting that countries with better social well-being and living standards also tend to have strong sustainability practices. Additionally, the Economic Index and Institutional Index exhibit a moderate correlation, implying that stronger economies are often supported by robust governance structures. 
The Year and Quality of Life sub-index show a moderate correlation, indicating that global living standards have generally improved over time. Similarly, the Economic and Quality of Life sub-indices are moderately correlated, which suggests that factors beyond economic strength, such as institutional quality and environmental sustainability, also significantly impact a country's quality of life.

\begin{figure}[htbp!]
\centering
\includegraphics[width=0.5\textwidth]{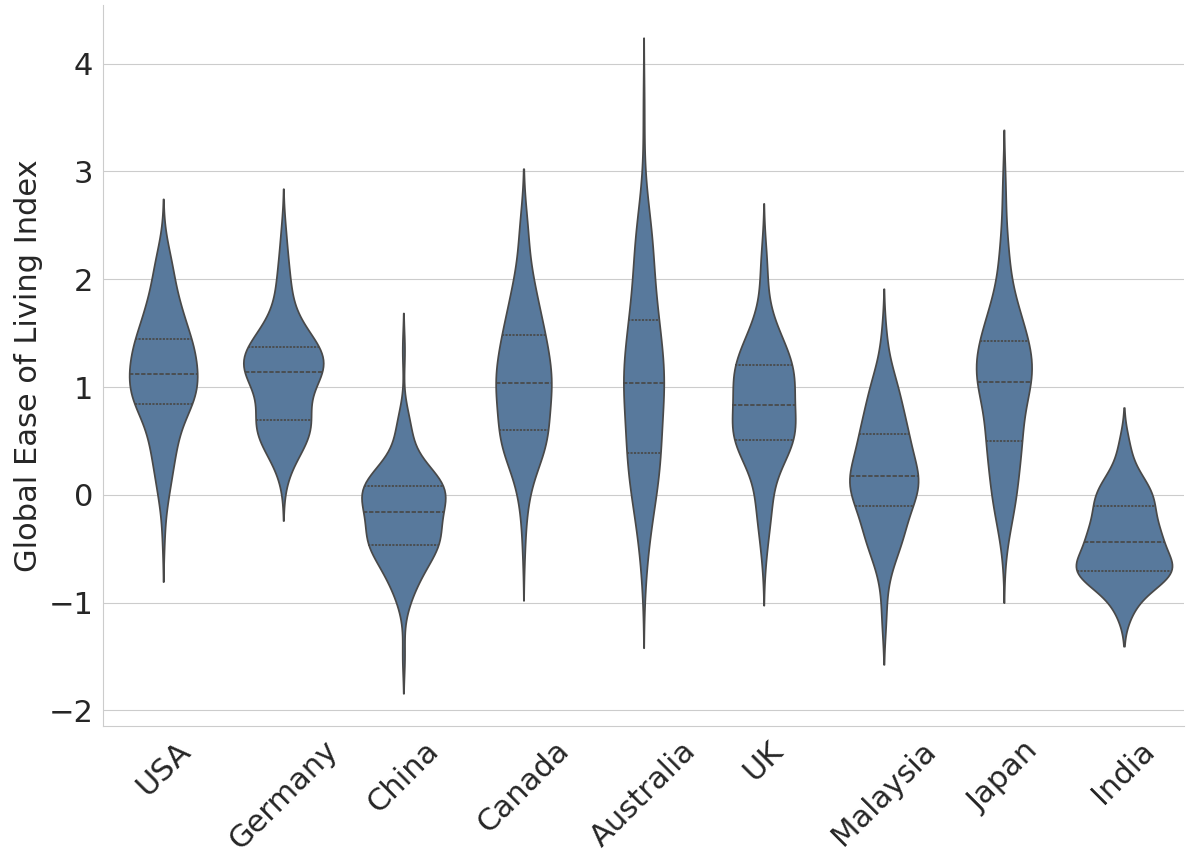}
\caption{The violin plot illustrates the distribution of the Global EoLI for selected countries over the last three decades, highlighting variations in data spread and central tendencies across regions.}
\label{fig:violin_plots}
\end{figure}

Figure \ref{fig:violin_plots} illustrates the violin plot of the distribution of our Global EoLI across the selected countries, highlighting both the range and density of the data.  We observe that the developed exhibit consistently high living standards, which can be deduced from their narrow violin and having higher values. Canada and Germany show stable, high median values with a bit more variability compared to the United States. In contrast, China, India, and Malaysia have lower median values, with India showing a particularly tight distribution around lower scores, reflecting relatively uniform, but lower living conditions, compared to more developed nations. This is natural since developing countries have a series of challenges, given resources, infrastructure, government services, and employment, which highly influence the quality of life. 

We further present a map-based visualisation of the selected countries in our study. 
The choropleth maps present the global distribution of the Ease of Living Index, representing the beginning of the given decades with years 1991, 2001, 2011 and 2021, respectively. We categorise the selected countries into four major levels, as shown in the Appendix.

We note that there has been media hype about the World Happiness Index \cite{helliwell2015geography}, which has attracted attention in social media and the government of countries such as India questioning the methodology and arguing that the rank should have been 48 rather than 126 in 2024 according to the SBI report \footnote{\url{https://timesofindia.indiatimes.com/india/world-happiness-index-flawed-indias-rank-should-have-been-48-not-126-sbi-ecowrap/articleshow/99337121.cms}}.
We review and compare the World Happiness Index with our Global Ease of Living Index. 

\begin{table*}[htbp!]
\centering
\small
\begin{tabular}{|l|c|c|c|c|c|c|}
\hline
\textbf{Country} & 
\textbf{Global EoLI} & 
\textbf{Economic} & 
\textbf{Institutional} & 
\textbf{Quality of Life} & 
\textbf{Sustainability} \\
\hline
Malaysia & \textbf{47}  & 53 & 38 & 54 & 193 \\
United States & 18  & 15 & 20 & 19 & 147 \\
China & \textbf{41} & 21 & 50 & 63 & 68 \\
Canada & 17  & 20 & 11 & 15 & 158 \\
Germany & 19  & 27 & 12 & 9 & 143 \\
Australia & 4 & 7 & 10 & 5 & 67 \\
India & \textbf{51}  & 29 & 57 & 75 & 118 \\
Japan & 8  & 6 & 15 & 18 & 95 \\
United Kingdom & 21  & 26 & 18 & 17 & 146 \\
\hline
\end{tabular}
\caption{Global EoLI and Sub-Index Ranks for 2021.}
\label{tab:happiness}
\end{table*}

Table \ref{tab:happiness} shows a comparison of Global EoLI rankings and Happiness Index rankings for 2021, showing notable disparities between the countries. For instance, the United States and Canada achieve relatively high rankings on both indices (18th and 17th for Ease of Living, 14th and 15th for Happiness, respectively), reflecting their strong economic foundations and extensive social welfare systems. In contrast, India displays a notable disparity: ranking 51st on the Global EoLI but plummeting to 139th on the Happiness Index. This gap is partly attributable to the Happiness Index’s emphasis on economic measures such as GDP per capita without considering the cost of living, which can unfairly disadvantage developing nations \cite{Akgün}. Moreover, the limited sampling by the World Happiness Index, which is based on collecting only 2,000 - 3,000 survey responses to represent large countries, such as India and China, that have populations exceeding 1.4 billion, fails to accurately reflect the complexity and diversity of various economic and social factors in different regions   \cite{Ko_Lee}. Malaysia similarly suffers from inadequate sample representation, which may skew results across its diverse regions. Furthermore, happiness outcomes across countries are strongly influenced by differences in ecological variables, such as access to natural resources, environmental quality, and income levels. In contrast, Japan ranks lower in happiness (56th) but has a high Global EoLI ranking (8th). This might be due to sociocultural factors, including ageing populations and work-related stress. This suggests that the World Happiness Index is a context-dependent metric that takes into account the particular socioeconomic and environmental characteristics of each nation rather than being a universal one \cite{Radkevich}. These limitations highlight the necessity for more representative, equitable and transparent data collection methods, and emphasise the importance of developing more nuanced indices that account for varying global contexts.

\begin{table*}[htbp!]
    \centering
    \caption{ \textcolor{black}{Average rankings of different sub-indices over the last decade}}
    \resizebox{\textwidth}{!}{%
    \begin{tabular}{|l|c|c|c|c|c|}
        \hline
        \textbf{Country} & \textbf{Economic Index} & \textbf{Institutional Index} & \textbf{Quality of Life Index} & \textbf{Sustainability Index} & \textbf{Ease of Living Index} \\
        \hline
        Australia & 7.4 & 10.1 & 7.5 & 6.0 & 4.2 \\
        Canada & 17.8 & 9.1 & 12.4 & 9.4 & 9.1 \\
        China & 37.6 & 63.3 & 71.8 & 37.6 & 52.9 \\
        Germany & 23.6 & 12.6 & 4.4 & 24.8 & 17.3 \\
        India & 49.7 & 58.9 & 80.0 & 77.5 & 66.4 \\
        Japan & 7.7 & 16.9 & 17.6 & 13.5 & 9.1 \\
        Malaysia & 56.7 & 39.8 & 54.6 & 19.2 & 46.4 \\
        United Kingdom & 17.4 & 13.4 & 15.1 & 29.8 & 16.1 \\
        United States & 16.1 & 18.3 & 15.9 & 18.0 & 12.8 \\
        \hline
    \end{tabular}%
    }

    \label{tab:average_rankings}
\end{table*}

We finally review the sub-indices and ranking for the selected countries for the last decade in relation to the Global EoLI. 
Table \ref{tab:average_rankings} shows significant differences in rankings across economic, institutional, quality of life, sustainability, and ease of living indices. Germany, Canada, and Australia continuously do well in every category, which is indicative of their strong economies, efficient political systems, excellent standards of living, and aggressive environmental programs. We notice that good governance and economic stability are demonstrated by Japan's good performance, especially in economic and institutional rankings. However, despite advancements in economic growth, China and India have higher average scores, indicating continued difficulties in enhancing sustainability and quality of life. Similar issues confront Malaysia, which performs poorly on sustainability and institutional indexes. In relation to their high economic and living standards, the United States and the United Kingdom have balanced but somewhat lower sustainability and institutional rankings. These insights emphasise the need for a multifaceted approach to national development that not only focuses on economic growth but also strengthens governance, social welfare, and environmental sustainability to improve overall ease of living.

\textcolor{black}{Figure \ref{fig:eoli2021} presents the Global EoLI for the selected countries from 1970-2024, which finds a close relationship with the Quality of Life sub-index as shown in Figure 5. Tables 8 and 9 in the Appendix present the names of the top 20 countries every 5 years for the Global EoLI from 1975. Further details given in the visualisation in Figures 19-22.}

\begin{figure*}[htbp!]
\centering
\includegraphics[width=0.79\textwidth]{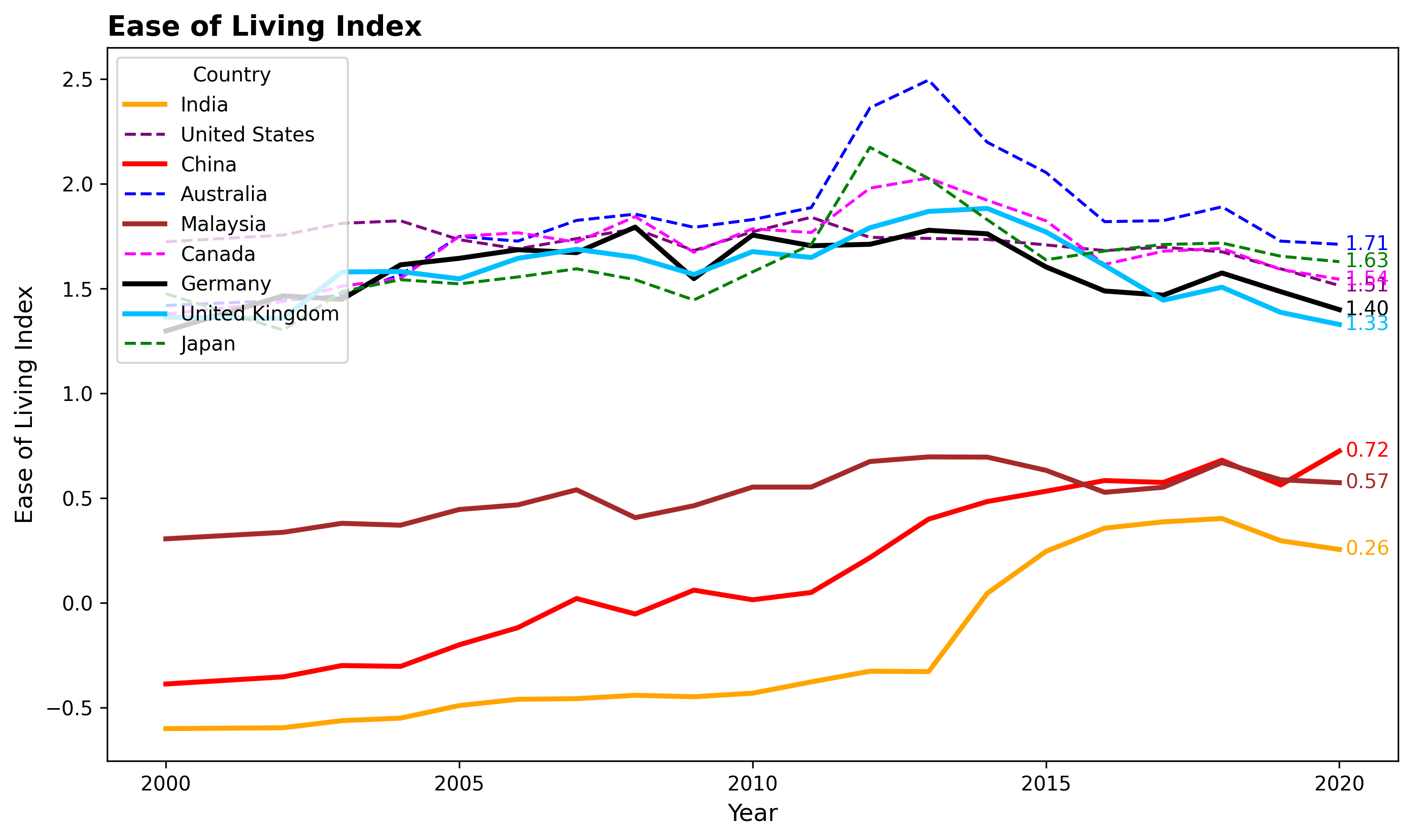}
\caption{Global Ease of Living Index}
\label{fig:eoli2021}
\end{figure*}

\subsection{Robustness analysis and comparison with established indices}

\begin{table*}[htbp]
\centering
\small
\caption{Country rankings under different weighting schemes for the four different sub-indices (Economic, Institutional, Quality of Life, and Sustainability).}
\label{tab:ranking_scenarios}
\begin{tabular}{lccccc}
\toprule
\textbf{Country} &
\textbf{Baseline} &
\textbf{Scenario 1} &
\textbf{Scenario 2} &
\textbf{Scenario 3} &
\textbf{Scenario 4} \\
&
(25,25,35,15) &
(20,30,30,20) &
(15,30,25,30) &
(25,25,25,25) &
(30,15,20,35) \\
\midrule
Australia      & 4  & 4  & 3  & 4  & 4  \\
Japan          & 8  & 7  & 5  & 5  & 3  \\
Canada         & 17 & 12 & 13 & 14 & 15 \\
United States  & 18 & 15 & 12 & 12 & 10 \\
Germany        & 19 & 14 & 11 & 16 & 16 \\
United Kingdom & 21 & 18 & 15 & 18 & 14 \\
China          & 41 & 37 & 40 & 33 & 29 \\
Malaysia       & 47 & 41 & 37 & 40 & 38 \\
India          & 51 & 47 & 52 & 45 & 42 \\
\bottomrule
\label{tab:diffweight}
\end{tabular}
\end{table*}

\textcolor{black}{We examine the robustness of the proposed Global EoLI  by comparing country rankings under alternative weighting schemes, as shown in Table~\ref{tab:ranking_scenarios}. The baseline weighting scheme assigns 25\% to the Economic sub-index, 25\% to the Institutional sub-index, 35\% to the Quality of Life sub-index, and 15\% to the Sustainability sub-index. This weighting approach is consistent with the construction of composite indicators, where multiple dimensions are aggregated using transparent and interpretable weighting schemes \cite{nardo2005handbook}. Across the alternative scenarios, the rankings of highly ranked countries such as Australia, Japan, Canada, the United States, Germany, and the United Kingdom remain relatively stable, suggesting that the Global-EoLI is not highly sensitive to moderate changes in sub-index weights. Australia remains within the top four across all weighting schemes, while Japan improves under scenarios that increase the importance of economic and sustainability-related dimensions. This indicates that the overall ranking structure is robust for countries with consistently strong performance across multiple dimensions.}

\textcolor{black}{The variation in rankings across the alternative weighting schemes can be explained by the fact that each country depends differently on the four sub-indices. Countries with strong performance in a particular dimension benefit when that dimension receives a higher weight, while countries with weaker performance in that area experience a decline in rank. For example, developed countries generally perform strongly in the Economic, Institutional, and Quality of Life sub-indices, reflecting mature governance systems, stable macroeconomic structures, stronger healthcare systems, and higher living standards. This is consistent with the broader literature showing that institutional quality and economic performance are central to long-run development outcomes \cite{north1990institutions,acemoglu2005institutions}. Therefore, when greater emphasis is placed on the institutional and economic dimensions, developed economies tend to retain or improve their rankings. In contrast, emerging economies such as China, Malaysia, and India show more noticeable rank changes because their performance is uneven across the four sub-indices. This demonstrates the importance of using a balanced multidimensional framework rather than relying on a single dominant component.}

\begin{table}[htbp]
\centering
\small 
\caption{Selected county rank by each Sub- including Quality of Life (QoL), Sustainability (Sus.), Economic (Eco.) and Institutional (Inst.). }
\label{tab:indicator_rankings}
\begin{tabular}{lcccc}
\toprule
\textbf{Country} &
\textbf{Eco.} &
\textbf{Inst.} &
\textbf{QoL} &
\textbf{Sus.} \\
\midrule
Australia      & 7  & 10 & 5  & 67 \\
Japan          & 6  & 15 & 18 & 95 \\
Canada         & 20 & 11 & 15 & 158 \\
USA  & 15 & 20 & 19 & 147 \\
Germany        & 27 & 12 & 9  & 143 \\
UK& 26 & 18 & 17 & 146 \\
China          & 21 & 50 & 63 & 68 \\
Malaysia       & 53 & 38 & 54 & 193 \\
India          & 29 & 57 & 75 & 118 \\
\bottomrule
\label{tab:diffindex}

\end{tabular}
\end{table}

\begin{table}[htbp!]
\centering
\small
\caption{Comparison of EoLI rank for G20 countries with selected global index ranks for 2021. The countries selected in our study for detailed analysis have been highlighted in bold.}
\label{tab:eoli_global_index_ranks}
\begin{tabular}{lcccccc}
\hline
Country & EoLI\tablefootnote{EoLI: Ease of Living Index Rank} 
& Gini\tablefootnote{Gini: Gini Index Rank} 
& PLI\tablefootnote{PLI: Price Level Index Rank} 
& CPI\tablefootnote{CPI: Consumer Price Index Rank} 
& NRR\tablefootnote{NRR: Natural Resources Rents Rank} 
& UHC\tablefootnote{UHC: Universal Health Coverage Service Coverage Rank} \\
\hline
\textbf{Australia}      & \textbf{4}  & \textbf{68} & \textbf{191} & \textbf{74}  & \textbf{146} & \textbf{3}  \\
\textbf{Japan}          & \textbf{8}  & \textbf{68} & \textbf{177} & \textbf{6}   & \textbf{24}  & \textbf{7}  \\
South Korea             & 12 & 26 & 159 & 47  & 25  & 5  \\
\textbf{Canada}         & \textbf{17} & \textbf{19} & \textbf{180} & \textbf{62}  & \textbf{111} & \textbf{1}  \\
\textbf{USA}  & \textbf{18} & \textbf{51} & \textbf{187} & \textbf{73}  & \textbf{75}  & \textbf{5}  \\
\textbf{Germany}        & \textbf{19} & \textbf{29} & \textbf{169} & \textbf{41}  & \textbf{26}  & \textbf{6}  \\
France                  & 20 & 21 & 167 & 19  & 19  & 10 \\
\textbf{UK} & \textbf{21} & \textbf{24} & \textbf{179} & \textbf{70}  & \textbf{58}  & \textbf{4}  \\
Italy                   & 28 & 37 & 161 & 20  & 31  & 10 \\
Saudi Arabia            & 37 & 68 & 118 & 83  & 165 & 10 \\
\textbf{China}          & \textbf{41} & \textbf{42} & \textbf{140} & \textbf{87}  & \textbf{82}  & \textbf{8}  \\
\textbf{India}          & \textbf{51} & \textbf{68} & \textbf{19}  & \textbf{140} & \textbf{100} & \textbf{25} \\
Argentina               & 56 & 58 & 84  & 178 & 96  & 12 \\
Turkey                  & 59 & 61 & 32  & 160 & 64  & 15 \\
South Africa            & 61 & 68 & 113 & 128 & 125 & 20 \\
Russia                  & 62 & 39 & 35  & 143 & 157 & 12 \\
Indonesia               & 67 & 41 & 43  & 119 & 113 & 26 \\
Mexico                  & 70 & 68 & 112 & 118 & 101 & 13 \\
Brazil                  & 83 & 66 & 101 & 139 & 130 & 8  \\
\hline
\end{tabular}
\end{table}

\textcolor{black}{Table~\ref{tab:indicator_rankings} provides further insight into the source of these rank changes by showing each selected country's rank across the four sub-indices. Australia performs strongly in the Economic, Institutional, and Quality of Life indices, which explains its stable position in Table~\ref{tab:ranking_scenarios}. Japan has a high Economic Index rank and a relatively strong Sustainability Index rank, supporting its improvement when economic and sustainability weights are increased. Canada and Germany perform well in the Institutional and Quality of Life dimensions, but their weaker Sustainability Index ranks limit their improvement when sustainability receives a larger weight. Similarly, China performs comparatively well in the Economic Index but ranks lower in the Institutional and Quality of Life dimensions, explaining its sensitivity to changes in the weighting structure. India shows a moderate Economic Index rank but weaker Institutional, Quality of Life, and Sustainability ranks, which explains its lower overall position under most weighting schemes.}

\textcolor{black}{Although sustainability is an important component of ease of living, assigning it an excessively high weight may not always provide a balanced interpretation of national living conditions. Sustainability improvements are often gradual, capital-intensive, and structurally different across developed and developing economies. In developed countries, the marginal cost of further sustainability improvement can be high because many basic environmental regulations and infrastructure transitions have already been implemented. In developing countries, sustainability progress may be constrained by industrialisation, energy demand, and infrastructure expansion \cite{munoz2020energy,chen2022environmental}. Therefore, giving a very high weight to the Sustainability Index alone could overstate one dimension of development and underrepresent the role of economic stability, institutional quality, and quality of life. The baseline weighting scheme (Table ~\ref{tab:ranking_scenarios}) is therefore appropriate because it recognises sustainability while still giving substantial importance to the economic, institutional, and social dimensions that directly shape ease of living.}

\textcolor{black}{Table~\ref{tab:eoli_global_index_ranks} compares the 2021 Global-EoLI rankings with selected global indicator ranks, including inequality, price levels, consumer prices, natural resource rents, and universal health coverage. The results show that the Global EoLI captures broader living conditions than any individual indicator. For instance, Australia, Japan, Canada, the United States, Germany, and the United Kingdom achieve strong Global-EoLI rankings despite variation in their price-level and consumer-price rankings, indicating that higher living costs in developed economies are partly offset by stronger healthcare coverage, institutional quality, and overall living standards. Conversely, India has a favourable Price Level Index rank, indicating relatively lower price levels, but its Global EoLI rank is lower due to weaker performance in other dimensions. Similarly, China performs moderately in several individual indicators but ranks lower than most developed countries in the Global EoLI, reflecting the multidimensional nature of the index. Overall, Table~\ref{tab:eoli_global_index_ranks} supports the argument that ease of living cannot be adequately assessed using isolated indicators; instead, it requires an integrated framework that combines economic, institutional, quality-of-life, and sustainability dimensions \cite{jones2016beyond,decancq2013copula}.}

\section{Discussion}


  Although MICE performed well as a statistical model when applied to structured data, it showed shortcomings when dealing with significant missing value unpredictability. Key themes emerged from analyses of each sub-index, with the Institutional sub-index showing how governance systems affect stability, the Quality of Life sub-index revealing differences between nations, and the Economic sub-index displaying macroeconomic stability and growth patterns. The Sustainability sub-index emphasized differing levels of environmental commitment, with developed nations leading in emissions reduction and renewable energy use.  Correlation analysis revealed a strong link between Quality of Life Index and Sustainability Index, comparisons with the Happiness Index revealed shortcomings of Happiness Index with regard to data representativeness and socioeconomic biases, especially in developing countries. This analysis underscores the diverse factors shaping global living conditions and the need for targeted, equitable improvements in quality of life.


Multiple imputation is widely regarded as a robust method for addressing missing data, as it allows for the creation of several complete datasets by imputing missing values based on observed data patterns \cite{Mutubuki_Alili}. This technique helps to preserve the statistical power of the analysis and reduces the bias that can arise from simpler methods, such as mean imputation, and last observation carried forward \cite{Alili_Dongen}.  Studies have implemented this approach for the imputation of socioeconomic and environmental data, demonstrating its effectiveness and flexibility. In alignment with these findings, we have utilised MICE, which employs multiple imputations across all sub-indices in our analysis. This reflects consistency with the established practices in the literature and underscores its suitability as a good statistical model for data imputation. RFR  demonstrated strong performance, even when up to 50\% of values are missing \cite{Kakaï}, with their effectiveness particularly noted in ecological data \cite{Kakaï}. According to a number of studies comparing different imputation strategies, models like missForest frequently offer strong performance across a range of missing data mechanisms  \cite{Miss_Forest}. According to our investigation, the RFR performed better than MICE, which is consistent with studies in the literature.  

We have taken cues from well-established approaches in the literature while creating our Global EoLI. In particular, we have used a weighted average approach to develop our score, which is in line with the four sub-pillars that the Indian government has established for their Ease of Living score \cite{mohua2018ease}. Furthermore, in our Quality of Life sub-index, we have incorporated various factors, encompassing standard of living, education, and health. This strategy draws inspiration from the  HDI \cite{undp_hdi}, one of the most well-known composite indices that assesses human development along three main dimensions: standard of living (Gross National Income per capita), education (mean and expected years of schooling), and health (life expectancy) \cite{undp_hdi}. We have utilised  PCA to capture latent variables that encapsulate the maximum amount of information from our data, resulting in composite scores for the respective sub-indices. A large body of research \cite{Srivastava,Delalić_Šikalo} showed the effectiveness of PCA in creating composite indexes, which has motivated our framework.  

Although our Global EoLI has yielded insightful information about living standards around the world thanks to thorough data analysis and rigorous methodology, there are still areas ripe for future enhancement. \textcolor{black}{Complex and incomplete data patterns could be handled much better by integrating sophisticated imputation techniques based on data augmentation \cite{khan2024review}
with Random Forest models. Furthermore, utilising Bayesian deep learning \cite{kulkarniChandra2024bayes} for imputation would provide an improved method of handling uncertainty, enhancing data precision and resilience in general. By integrating Bayesian ranking techniques \cite{rendle2012bpr} into index creation, comparative analysis can be improved, and a fair rank can be established.}

In terms of limitations, we note that data imputation methods suffer from bias depending on the nature of the missing data, and whether it is missing at random. Therefore, effective methods need to be developed so that uncertainty can be quantified, allowing Bayesian inference  \cite{
kong1994sequential,chen2024deep} to provide a natural way to quantify uncertainty in data imputation models. Furthermore, we also note that the quality of the data can also be a problem, such as under-reporting of crimes in developing countries, and the limitations in the methodology for the choice of index construction.  \textcolor{black}{PCA, for example, assumes a correlation between features, but some of these features may not be correlated. Additionally, it is not robust against outliers in the data. We provide open-source code implementation for the extension of this list in the coming years, which can guide policymakers.} 
\textcolor{black}{Another limitation of this study relates to the compilation of the dataset from multiple sources, including international organisations and external databases. Although this approach is appropriate given the broad scope and cross-country nature of the analysis, it introduces potential challenges associated with data harmonisation. Differences in measurement methodologies, reporting standards, temporal coverage, and data collection practices across countries may affect the comparability and consistency of the resulting dataset. In addition, variations in data quality and completeness across sources could introduce uncertainty into the analysis and potentially influence cross-country comparisons.}
 
\textcolor{black}{Although indices such as the Human Development Index and the World Happiness Report provide valuable cross-sectional insights, they often rely on static weighting schemes and limited temporal coverage. Additionally, missing data challenges are typically handled through simple interpolation or listwise deletion, potentially introducing bias in longitudinal comparisons.}\textcolor{black}{ A key limitation of the Human Development Index dataset is the high proportion of missing observations for several indicators, including the Cost of Living Index (87.58\%), Healthcare Index (92.44\%), and Crime Index (88.78\%) for specific periods, as shown in  Tables 13-16 in the Appendix for the different indices.  Although machine learning–based imputation methods such as Random Forest Regressor and  MICE  were used to estimate the missing values, the reliability of indicators with extremely high missingness remains uncertain. The imputation models were evaluated using reconstruction experiments with metrics such as RMSE, and  KDE comparisons (Figure \ref{fig:econkde}) suggest that the imputed values broadly preserve the distributional characteristics of the observed data. However, these approaches cannot fully guarantee the accuracy of individual imputed observations.  Variables with extremely high levels of missingness rely heavily on model-based reconstruction and should therefore be interpreted with caution.
}

In future work, broadening the scope of parameters within the Sustainability Index—by incorporating metrics such as biodiversity, waste management practices, and climate resilience—would yield a more comprehensive view of environmental and societal health. Furthermore, expanding empirical validation through extensive physical surveys, for example, conducting surveys in major Australian cities, could improve our understanding of living situations and strengthen data reliability. Creating an interactive dashboard for Global EoLI  would enable users to interact with the index in real-time, promoting dynamic data additions. These efforts, when combined, will better guarantee that the index stays relevant and flexible for the changing global conditions, encouraging a group effort toward sustainable growth and improved living standards. Finally, effective strategies to quantify uncertainties by incorporating Bayesian inference in our framework can provide further improvements. \textcolor{black}{Our framework features major components and steps involving data imputation, dimensionality reduction, weighting, and aggregatio and hence the resulting index could be sensitive to modelling choices. In future work, it be valuable to include robustness checks that includes, bootstrap analysis, Monte Carlo simulations, and sensitivity tests for the weighting scheme and input variables.}

\section{Conclusion}
 
We presented the Global Ease of Living Index to reflect the complex relationship between living conditions in selected countries. We used machine learning techniques for data imputation, notably MICE and Random Forest models. The Random Forest imputer provided a more robust and dependable strategy for dealing with missing information. We constructed four sub-indices, including Economic, Institutional, Quality of Life, and Sustainability, using PCA and Factor Analysis,  contributing to a holistic understanding of global living conditions. The analysis revealed stark contrasts between developed economies, such as Australia, the United States, Japan and Germany, which consistently rank high due to their stable institutions and economic prosperity, and emerging nations such as India, China and Malaysia, which demonstrated impressive economic growth but continue to face challenges in social and environmental areas.

 Additionally, our comparison with the notable Happiness Index revealed notable biases, particularly the index's excessive dependence on GDP and small sample size based on surveys that fail to effectively reflect large populations in  China and India. This exposes the limitations of using a one-size-fits-all approach to measure well-being. Our index, on the other hand, acknowledges the context-dependent nature of quality of life by using a more diverse range of indicators. Our research highlights the value of reproducible and transparent data, providing policymakers with a useful framework to pinpoint areas that need improvement, ranging from environmental sustainability to healthcare access.  Therefore, this research paves the way for more nuanced and equitable strategies for enhancing global living standards, encouraging ongoing research and policy innovation to address the complexities of modern well-being.

\section*{Data and Code Avaialbaility}

Code and data: \url{https://github.com/pinglainstitute/ease-of-living-index}

\section*{Conflict of Interest}

The authors declare no conflict of interest exists. 




 \bibliographystyle{elsarticle-num} 
 \bibliography{cas-refs}




\clearpage
\appendix
\subsection{Map: Global Ease of Living Index}

\begin{figure*}[htbp!]
\centering
\includegraphics[width=0.9\textwidth]{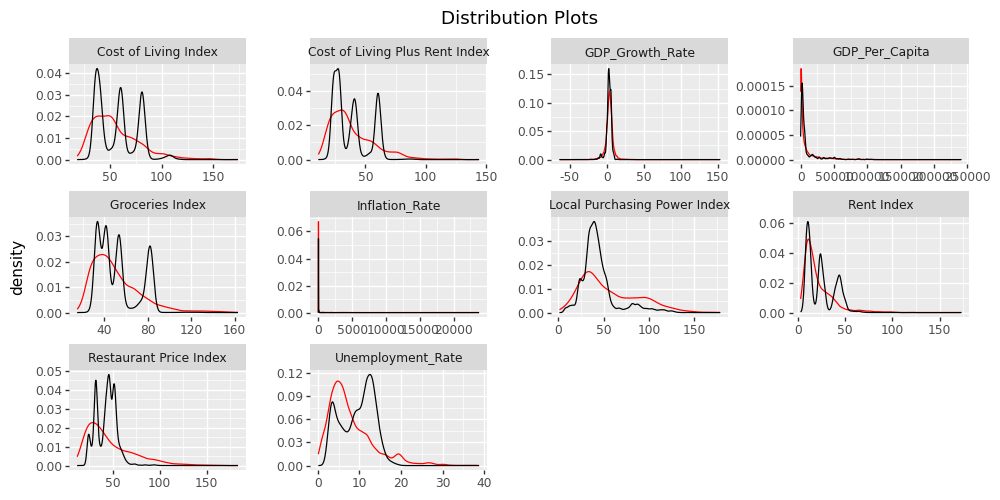}
\caption{KDE Plots for sub-indices showing Economic indicators. The Black curve represents the distribution of the original data, while the red curves depict the distribution of the data that has been imputed using MICE. } 
\label{fig:econkde}
\end{figure*}

Figure \ref{fig:econkde} \textcolor{black}{shows panel compares empirical densities of ten economic indicators, showing that modal locations and central tendencies align closely across most measures while spreads and tail behaviour are largely preserved. Minor smoothing and small dispersion shifts appear in a few panels (notably GDP per capita and Local Purchasing Power) but do not materially change substantive conclusions. Overall, the figure indicates the alternative series maintains the original distributions sufficiently to support subsequent comparative and aggregate analyses.}



\begin{figure*}[htbp!]
\centering
\includegraphics[width=0.9\textwidth]{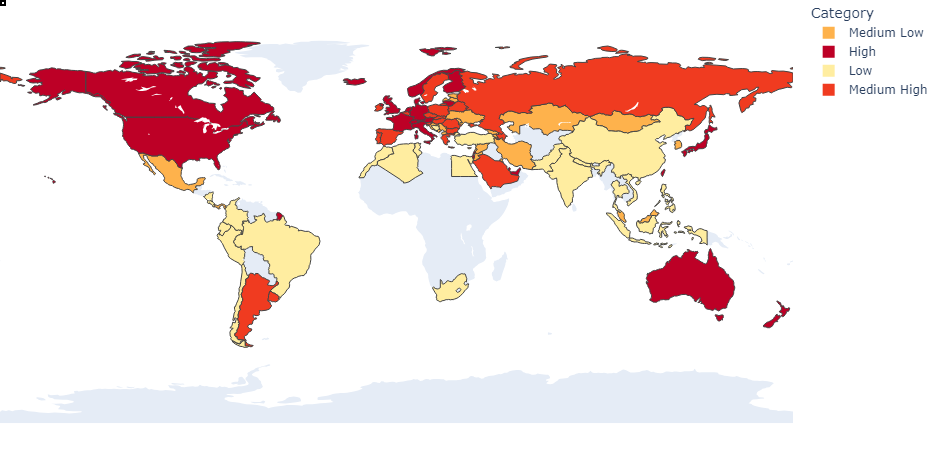}
\caption{Global Distribution of Ease of Living Index for the Year 1991}
\label{fig:eoli1991}
\end{figure*}

\begin{figure*}[htbp!]
\centering
\includegraphics[width=0.9\textwidth]{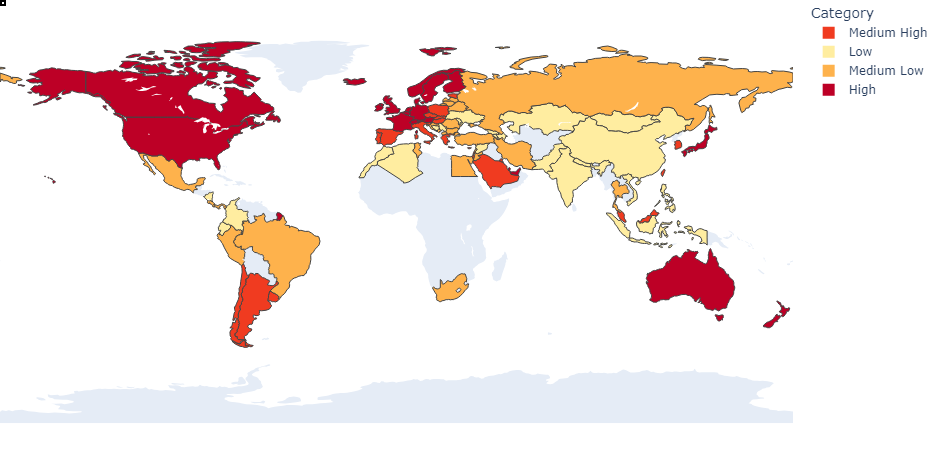}
\caption{Global Distribution of Ease of Living Index for the Year 2000}
\label{fig:eoli2000}
\end{figure*}

Figure~\ref{fig:eoli1991} (1991) and Figure~\ref{fig:eoli2000} (2000) highlight significant disparities in living standards, with many countries in Africa, South Asia, and Latin America classified as Low or Medium Low. In contrast, regions such as  North America, Western Europe, and Australia are predominantly categorised as High, reflecting well-established infrastructure, strong governance, and superior quality of life in these developed areas.

\begin{figure*}[htbp!]
\centering
\includegraphics[width=0.9\textwidth]{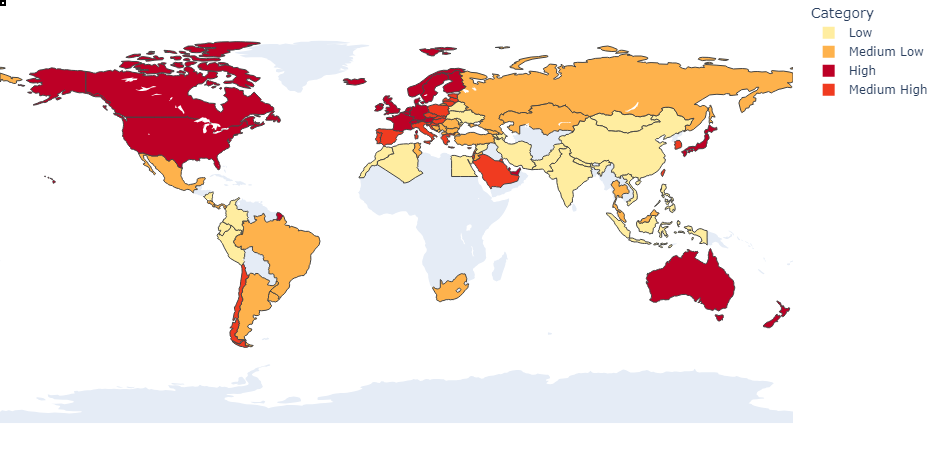}
\caption{Global Distribution of Ease of Living Index for the Year 2011}
\label{fig:eoli2011}
\end{figure*}

Figure~\ref{fig:eoli2021} (2021) shows notable improvements from Figure~\ref{fig:eoli2011} (2011), particularly in East Asia, the Middle East, and parts of Eastern Europe, where countries have shifted from lower to higher categories. This progress underscores the impact of economic growth, improved governance, and infrastructure development. Nevertheless, regions such as sub-Saharan Africa and parts of South Asia continue to struggle with lower scores, highlighting persistent challenges in achieving widespread development. Furthermore, countries such as India and China can be seen improving throughout all 4 maps. Overall, comparison across different years, from 1991 through 2021, reveals both progress and persistent inequality. While many emerging economies have experienced advancements, developed regions still maintain a significant lead in living standards. These findings emphasise the ongoing need for targeted development strategies to reduce global disparities and promote sustainable improvements in living conditions worldwide. Tables 11 and 12 present the top 20 countries by Global EoLi from 1975 to 2020.


\clearpage

\vspace{0.02\textheight}

\clearpage

\vspace{0.02\textheight}

\begin{figure*}[htbp!]
\centering
\includegraphics[width=0.9\textwidth]{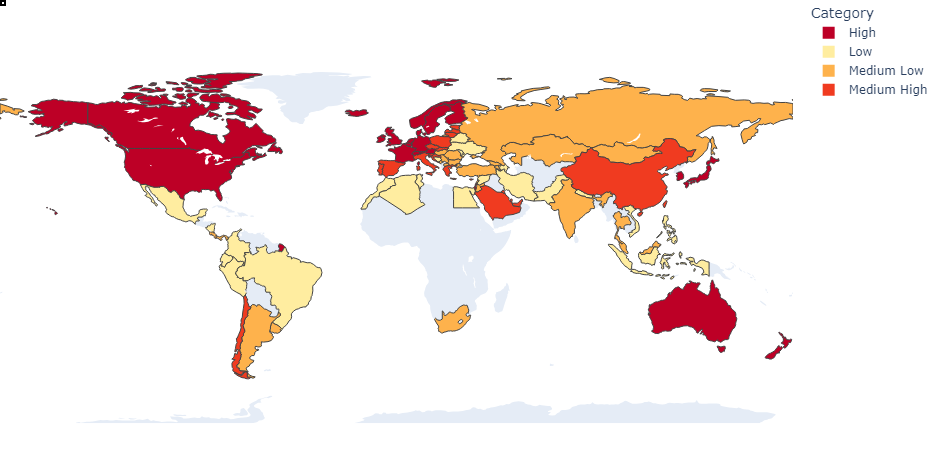}
\caption{Global Distribution of Ease of Living Index for the Year 2021}
\label{fig:eoli2021}
\end{figure*}

\begin{table*}[htbp!]
\centering
\small
\caption{ Top 20 Countries by Ease of Living Index (1975 -1995)}

\resizebox{\textwidth}{!}{%
\begin{tabular}{|c|c|c|c|c|c|}
\hline

Rank & 1975 & 1980 & 1985 & 1990 & 1995 \\
\hline

1 & Switzerland & France & Iceland & Switzerland & Denmark \\
2 & Sweden & Netherlands & Norway & Cayman Islands & United States \\
3 & Denmark & Denmark & Canada & United States & Switzerland \\
4 & Qatar & Germany & Sweden & Liechtenstein & Norway \\
5 & Puerto Rico & Bermuda & Switzerland & Denmark & Netherlands \\
6 & Euro area & Slovenia & Netherlands & Germany & Germany \\
7 & Bermuda & Switzerland & Australia & Norway & Japan \\
8 & Norway & Puerto Rico & Belgium & Australia & Australia \\
9 & Australia & Norway & United States & Netherlands & Austria \\
10 & Iceland & Belgium & European Union & Canada & Belgium \\
11 & Poland & United States & Kuwait & Japan & Canada \\
12 & Belgium & Austria & Qatar & Finland & Finland \\
13 & North America & Sweden & Finland & Austria & United Kingdom \\
14 & Canada & Qatar & France & Sweden & Sweden \\
15 & United Kingdom & United Kingdom & Bermuda & Iceland & United Arab Emirates \\
16 & Lithuania & Canada & Post-demographic dividend & Israel & Singapore \\
17 & Sint Maarten (Dutch part) & Kuwait & Puerto Rico & New Zealand & New Zealand \\
18 & United States & United Arab Emirates & OECD members & European Union & France \\
19 & Finland & Poland & Denmark & Belgium & Iceland \\
20 & Bulgaria & Euro area & Japan & Faroe Islands & Qatar \\

\hline
\end{tabular}%
}

\end{table*}

\begin{table*}[htbp!]
\centering
\small
\caption{ Top 20 Countries by Ease of Living Index (2000 - 2020)}

\resizebox{\textwidth}{!}{%
\begin{tabular}{|c|c|c|c|c|c|}
\hline

Rank & 2000 & 2005 & 2010 & 2015 & 2020 \\
\hline

1 & United States & Canada & Norway & Switzerland & Switzerland \\
2 & Switzerland & Australia & Switzerland & Norway & Norway \\
3 & Norway & United States & Denmark & Australia & Singapore \\
4 & Qatar & Switzerland & Ireland & Singapore & Australia \\
5 & Denmark & Norway & Australia & Denmark & Iceland \\
6 & Netherlands & Finland & Canada & New Zealand & Japan \\
7 & Japan & Netherlands & United States & Canada & Denmark \\
8 & Australia & Denmark & Germany & Ireland & Netherlands \\
9 & Canada & Germany & Netherlands & United Kingdom & Ireland \\
10 & United Kingdom & Ireland & Finland & Finland & Canada \\
11 & Finland & Austria & United Kingdom & Netherlands & New Zealand \\
12 & Sweden & Sweden & Sweden & Iceland & United States \\
13 & Germany & Belgium & Singapore & United States & Finland \\
14 & Ireland & New Zealand & Qatar & Qatar & Qatar \\
15 & Belgium & United Kingdom & Belgium & Belgium & Sweden \\
16 & Singapore & Qatar & New Zealand & Japan & Germany \\
17 & Iceland & Japan & Japan & Sweden & Belgium \\
18 & Austria & Iceland & Austria & Germany & United Arab Emirates \\
19 & New Zealand & United Arab Emirates & France & France & Austria \\
20 & France & France & United Arab Emirates & Austria & Israel \\

\hline
\end{tabular}%
}

\label{tab:ease_living_rankings}
\end{table*}
  
\clearpage
\subsection{Missing Values}

 \begin{table}[htbp]
\centering
\small
\renewcommand{\arraystretch}{1.05}
\caption{Summary of Missingness in the Economic Index Dataset}
\label{tab:economic_missingness}

\begin{tabular}{llcc}
\toprule
\textbf{Year Group} & \textbf{Missingness Category} & \textbf{Count} & \textbf{\%} \\
\midrule

\multicolumn{4}{c}{\textit{Overall Summary}} \\
\midrule
All Years & Complete & 1181 & 9.20 \\
& Low (1 variable) & 41 & 0.32 \\
& Moderate (2--3 variables) & 9368 & 72.99 \\
& High (4--5 variables) & 2245 & 17.49 \\

\midrule
\multicolumn{4}{c}{\textit{Year-wise Summary}} \\
\midrule

\multicolumn{4}{l}{\textbf{1970--1979} \hfill Avg Missing Index: 57.43\%} \\
& Moderate (2--3 variables) & 1091 & 58.62 \\
& High (4--5 variables) & 770 & 41.38 \\

\multicolumn{4}{l}{\textbf{1980--1989} \hfill Avg Missing Index: 55.34\%} \\
& Moderate (2--3 variables) & 1490 & 70.45 \\
& High (4--5 variables) & 625 & 29.55 \\

\multicolumn{4}{l}{\textbf{1990--1999} \hfill Avg Missing Index: 39.40\%} \\
& Moderate (2--3 variables) & 2244 & 92.69 \\
& High (4--5 variables) & 177 & 7.31 \\

\multicolumn{4}{l}{\textbf{2000--2009} \hfill Avg Missing Index: 36.73\%} \\
& Moderate (2--3 variables) & 2383 & 94.83 \\
& High (4--5 variables) & 130 & 5.17 \\

\multicolumn{4}{l}{\textbf{2010--2019} \hfill Avg Missing Index: 28.45\%} \\
& Complete & 759 & 28.19 \\
& Low (1 variable) & 12 & 0.45 \\
& Moderate (2--3 variables) & 1670 & 62.04 \\
& High (4--5 variables) & 251 & 9.32 \\

\multicolumn{4}{l}{\textbf{2020--2029} \hfill Avg Missing Index: 30.66\%} \\
& Complete & 422 & 34.23 \\
& Low (1 variable) & 29 & 2.35 \\
& Moderate (2--3 variables) & 490 & 39.74 \\
& High (4--5 variables) & 292 & 23.68 \\

\bottomrule
\end{tabular}
\end{table}

\begin{table}[htbp!]
\centering
\small
\caption{Summary of missingness for Quality of Life Index Dataset} 
\begin{tabular}{llrr}
\hline
\textbf{Year Group} & \textbf{Category} & \textbf{Count} & \textbf{\%} \\
\hline
\multicolumn{4}{c}{\textit{Overall Summary}} \\
\hline
All Years & Complete & 563 & 4.14 \\
All Years & Low Missingness (1--2 variables) & 2384 & 17.53 \\
All Years & Moderate Missingness (3--4 variables) & 2058 & 15.13 \\
All Years & High Missingness (5--6 variables) & 4107 & 30.20 \\
All Years & Very High Missingness (6+ variables) & 4486 & 32.99 \\
\hline
\multicolumn{4}{c}{\textit{By Year Group}} \\
\hline
\multicolumn{4}{l}{\textbf{1970--1979 \hfill Avg Missing Index: 85.33\%}} \\
1970--1979 & High Missingness (5--6 variables) & 440 & 17.36 \\
1970--1979 & Very High Missingness (6+ variables) & 2094 & 82.64 \\
\multicolumn{4}{l}{\textbf{1980--1989 \hfill Avg Missing Index: 84.37\%}} \\
1980--1989 & High Missingness (5--6 variables) & 639 & 25.02 \\
1980--1989 & Very High Missingness (6+ variables) & 1915 & 74.98 \\
\multicolumn{4}{l}{\textbf{1990--1999 \hfill Avg Missing Index: 56.64\%}} \\
1990--1999 & Low Missingness (1--2 variables) & 463 & 17.96 \\
1990--1999 & Moderate Missingness (3--4 variables) & 836 & 32.43 \\
1990--1999 & High Missingness (5--6 variables) & 841 & 32.62 \\
1990--1999 & Very High Missingness (6+ variables) & 438 & 16.99 \\
\multicolumn{4}{l}{\textbf{2000--2009 \hfill Avg Missing Index: 46.97\%}} \\
2000--2009 & Low Missingness (1--2 variables) & 912 & 35.32 \\
2000--2009 & Moderate Missingness (3--4 variables) & 698 & 27.03 \\
2000--2009 & High Missingness (5--6 variables) & 946 & 36.64 \\
2000--2009 & Very High Missingness (6+ variables) & 26 & 1.01 \\
\multicolumn{4}{l}{\textbf{2010--2019 \hfill Avg Missing Index: 37.52\%}} \\
2010--2019 & Complete & 474 & 18.39 \\
2010--2019 & Low Missingness (1--2 variables) & 826 & 32.04 \\
2010--2019 & Moderate Missingness (3--4 variables) & 372 & 14.43 \\
2010--2019 & High Missingness (5--6 variables) & 896 & 34.76 \\
2010--2019 & Very High Missingness (6+ variables) & 10 & 0.39 \\
\multicolumn{4}{l}{\textbf{2020--2029 \hfill Avg Missing Index: 46.00\%}} \\
2020--2029 & Complete & 89 & 11.53 \\
2020--2029 & Low Missingness (1--2 variables) & 183 & 23.70 \\
2020--2029 & Moderate Missingness (3--4 variables) & 152 & 19.69 \\
2020--2029 & High Missingness (5--6 variables) & 345 & 44.69 \\
2020--2029 & Very High Missingness (6+ variables) & 3 & 0.39 \\
\hline
\end{tabular}
\label{tab:qol_missingness}
\end{table}

 \begin{table}[htbp]
\centering
\small
\renewcommand{\arraystretch}{1.03}
\caption{Summary of Missingness in the Sustainability Index Dataset}
\label{tab:sustainability_missingness}

\begin{tabular}{llcc}
\toprule
\textbf{Year Group} & \textbf{Missingness Category} & \textbf{Count} & \textbf{\%} \\
\midrule

\multicolumn{4}{c}{\textit{Overall Summary}} \\
\midrule
All Years & Complete & 3377 & 23.93 \\
& Low (1 variable) & 1150 & 8.15 \\
& Moderate (2--3 variables) & 6686 & 47.38 \\
& High (4 variables) & 262 & 1.86 \\
& Fully Missing & 2636 & 18.68 \\

\midrule
\multicolumn{4}{c}{\textit{Year-wise Summary}} \\
\midrule

\multicolumn{4}{l}{\textbf{1970--1979} \hfill Avg Missing Index: 76.74\%} \\
& Moderate (2--3 variables) & 1393 & 57.90 \\
& High (4 variables) & 12 & 0.50 \\
& Fully Missing & 1001 & 41.60 \\

\multicolumn{4}{l}{\textbf{1980--1989} \hfill Avg Missing Index: 76.48\%} \\
& Moderate (2--3 variables) & 1564 & 58.77 \\
& High (4 variables) & 1 & 0.04 \\
& Fully Missing & 1096 & 41.19 \\

\multicolumn{4}{l}{\textbf{1990--1999} \hfill Avg Missing Index: 24.16\%} \\
& Complete & 1201 & 45.15 \\
& Low (1 variable) & 414 & 15.56 \\
& Moderate (2--3 variables) & 831 & 31.24 \\
& High (4 variables) & 123 & 4.62 \\
& Fully Missing & 91 & 3.42 \\

\multicolumn{4}{l}{\textbf{2000--2009} \hfill Avg Missing Index: 19.77\%} \\
& Complete & 1360 & 51.13 \\
& Low (1 variable) & 460 & 17.29 \\
& Moderate (2--3 variables) & 715 & 26.88 \\
& High (4 variables) & 55 & 2.07 \\
& Fully Missing & 70 & 2.63 \\

\multicolumn{4}{l}{\textbf{2010--2019} \hfill Avg Missing Index: 30.77\%} \\
& Complete & 816 & 30.68 \\
& Low (1 variable) & 276 & 10.38 \\
& Moderate (2--3 variables) & 1440 & 54.14 \\
& High (4 variables) & 54 & 2.03 \\
& Fully Missing & 74 & 2.78 \\

\multicolumn{4}{l}{\textbf{2020--2029} \hfill Avg Missing Index: 68.20\%} \\
& Moderate (2--3 variables) & 743 & 69.83 \\
& High (4 variables) & 17 & 1.60 \\
& Fully Missing & 304 & 28.57 \\

\bottomrule
\end{tabular}
\end{table}

\begin{table}[htbp!]
\centering
\caption{Summary of missingness for Institutional Index Dataset}

\begin{tabular}{llrr}
\hline
\textbf{Year Group} & \textbf{Category} & \textbf{Count} & \textbf{Percentage (\%)} \\
\hline
\multicolumn{4}{c}{\textit{Overall Summary}} \\
\hline
All Years & Complete & 5136 & 100.00 \\
\hline
\multicolumn{4}{c}{\textit{By Year Group}} \\
\hline
\multicolumn{4}{l}{\textbf{1990--1999 \hfill Avg Missing Index: 0.00\%}} \\
1990--1999 & Complete & 428 & 100.00 \\
\multicolumn{4}{l}{\textbf{2000--2009 \hfill Avg Missing Index: 0.00\%}} \\
2000--2009 & Complete & 1926 & 100.00 \\
\multicolumn{4}{l}{\textbf{2010--2019 \hfill Avg Missing Index: 0.00\%}} \\
2010--2019 & Complete & 2140 & 100.00 \\
\multicolumn{4}{l}{\textbf{2020--2029 \hfill Avg Missing Index: 0.00\%}} \\
2020--2029 & Complete & 642 & 100.00 \\
\hline
\end{tabular}
\label{tab:institutional_missingness}
\end{table}

The missingness patterns across the four sub-index datasets are summarised in Tables~\ref{tab:economic_missingness}--\ref{tab:sustainability_missingness}. 
Table~\ref{tab:economic_missingness} presents the missingness structure of the Economic Index dataset. Overall, most observations fall under the moderate missingness category, with 9368 observations, or 72.99\%, missing between two and three variables. Only 9.20\% of the observations are fully complete, while 17.49\% fall under high missingness. The decade-wise summary shows that missingness was highest in the earlier periods, with an average missing index of 57.43\% during 1970--1979 and 55.34\% during 1980--1989. However, data completeness improves over time, particularly after 2010, where the proportion of complete observations increases to 28.19\% in 2010--2019 and 34.23\% in 2020--2029. This indicates that the Economic Index dataset has stronger reliability in recent decades, although imputation remains necessary for earlier periods.

Table~\ref{tab:qol_missingness} reports the missingness summary for the Quality of Life Index dataset. Compared with the Economic Index, this dataset shows a more severe missingness structure. Across all years, only 4.14\% of observations are complete, while 30.20\% fall under high missingness and 32.99\% fall under very high missingness. The decade-wise results show that the earliest periods have the greatest data gaps, with average missing indices of 85.33\% in 1970--1979 and 84.37\% in 1980--1989. Missingness declines substantially from 1990 onward, reaching 37.52\% in 2010--2019. Although the 2020--2029 period shows a slight increase to 46.00\%, the proportion of very high missingness remains very low. This suggests that the Quality of Life indicators have limited historical coverage, but the dataset becomes more informative in recent decades.

Table~\ref{tab:institutional_missingness} shows the missingness summary for the Institutional Index dataset. Unlike the other sub-indices, the Institutional dataset is fully complete across all available observations. All 5136 observations are recorded as complete, corresponding to 100.00\% completeness across all years. The decade-wise breakdown also confirms 100.00\% completeness for each period from 1990--1999 to 2020--2029, with an average missing index of 0.00\%. Therefore, no imputation is required for the Institutional Index dataset, and the resulting institutional sub-index is not affected by missing-value reconstruction.

Table~\ref{tab:sustainability_missingness} summarises the missingness structure of the Sustainability Index dataset. Overall, 23.93\% of observations are complete, while the largest proportion, 47.38\%, falls under moderate missingness. However, 18.68\% of observations are fully missing, indicating that the Sustainability dataset contains more uneven data availability than the Economic and Institutional datasets. The decade-wise results show high missingness in earlier periods, with average missing indices of 76.74\% in 1970--1979 and 76.48\% in 1980--1989. Data availability improves considerably during 1990--1999 and 2000--2009, where complete observations increase to 45.15\% and 51.13\%, respectively. However, missingness increases again in 2020--2029, with an average missing index of 68.20\%. This pattern suggests that the Sustainability Index contains heterogeneous indicators with uneven temporal coverage, and therefore its results should be interpreted with greater caution than the other sub-indices.
 
 \newpage
\subsection{Source of data}

\begin{table}[htbp]
\centering
\small
\renewcommand{\arraystretch}{1.25}
\small
\caption{Data Sources for the Global EoLI}
\label{tab:data_sources}

\begin{tabular}{p{6.3cm} p{9.2cm}}
\toprule
\textbf{Factor} & \textbf{Data Source} \\
\midrule

Access to electricity &
\url{https://data.worldbank.org/indicator/EG.ELC.ACCS.ZS} \\

Annual healthcare expenditure &
\url{https://apps.who.int/nha/database} \\

Cost of Living Index &
\url{https://www.numbeo.com/cost-of-living/rankings_by_country.jsp} \\

Crime Index &
\url{https://www.numbeo.com/crime/rankings_by_country.jsp} \\

Corruption Perception Index &
\url{https://www.transparency.org/en/cpi/2021} \\

Ease of Doing Business Index &
\url{https://archive.doingbusiness.org/en/rankings} \\

Freedom of Speech &
\url{https://worldbank.org} \\

GDP Based on Purchasing Power Parity (PPP) &
\url{https://www.imf.org/external/datamapper/PPPSH@WEO/OEMDC/ADVEC/WEOWORLD?year=2024} \\

GDP per Capita &
\url{https://www.imf.org/external/datamapper/NGDPDPC@WEO/OEMDC/ADVEC/WEOWORLD} \\

Global Terrorism Index &
\url{https://prosperitydata360.worldbank.org/en/indicator/QOG+BD+voh_gti} \\

Healthcare Index &
\url{https://www.numbeo.com/health-care/rankings_by_country.jsp} \\

Inflation Rate (Average Consumer Prices) &
\url{https://www.imf.org/external/datamapper/PCPIPCH@WEO/WEOWORLD?year=2024} \\

Life expectancy &
\url{https://data.worldbank.org/indicator/SP.DYN.LE00.IN} \\

MYSE Index &
\url{https://genderdata.worldbank.org/en/indicator/se-sch-life} \\

Population Living in Slums &
\url{https://data.unhabitat.org/pages/housing-slums-and-informal-settlements} \\

Property Price-to-Income Ratio &
\url{https://www.numbeo.com/property-investment/rankings_by_country.jsp} \\

Real GDP Growth &
\url{https://www.imf.org/external/datamapper/NGDP_RPCH@WEO/WEOWORLD?year=2024} \\

Rule of Law Index &
\url{https://worldjusticeproject.org/rule-of-law-index/} \\

Unemployment Rate (ILO Data) &
\url{https://ilostat.ilo.org/data/\#} \\

Unemployment Rate (World Bank Data) &
\url{https://data.worldbank.org/indicator/SL.UEM.TOTL.ZS} \\

\bottomrule
\end{tabular}

\end{table}
 
\end{document}